\documentclass[10pt,twocolumn,letterpaper]{article}

\usepackage{cvpr}              % To produce the CAMERA-READY version
\usepackage[title]{appendix}

\usepackage{graphicx}
\usepackage{amsmath}
\usepackage{amssymb}
\usepackage{booktabs}
\usepackage{pifont}
\usepackage{multirow}
\usepackage{microtype}      % microtypography
\usepackage[table,dvipsnames]{xcolor}         % colors

\usepackage[inline]{enumitem}

\newenvironment{tight_itemize}{
\begin{itemize}[leftmargin=15pt,nosep]
  \setlength{\topsep}{0pt}
  \setlength{\itemsep}{0pt}
  \setlength{\parskip}{0pt}
  \setlength{\parsep}{0pt}
}{\end{itemize}}

\usepackage[pagebackref,breaklinks,colorlinks]{hyperref}

\newcommand{\GeoI}{GeoImNet}
\newcommand{\GeoP}{GeoPlaces}
\newcommand{\GeoU}{GeoUniDA}
\newcommand{\Ours}{GeoNet}

\newcommand{\rfy}[1]{Resnet50}
\newcommand{\vits}[1]{ViT-S}
\newcommand{\vitb}[1]{ViT-B}
\newcommand{\vitl}[1]{ViT-L}

\newcommand{\MYhref}[3][blue]{\href{#2}{\color{#1}{#3}}}%

\newcommand{\supp}{\textcolor{black}{supplementary material}}
\newcommand{\review}[1]{#1}

\newcommand{\tk}[1]{#1}
\newcommand{\noop}[1]{#1}

\DeclareCaptionFont{xipt}{\fontsize{8}{10}\mdseries}
\usepackage[font=xipt]{caption}

\usepackage[capitalize]{cleveref}
\crefname{section}{Sec.}{Secs.}
\Crefname{section}{Section}{Sections}
\Crefname{table}{Table}{Tables}
\crefname{table}{Tab.}{Tabs.}

\begin{document}

\title{\Ours{}: Benchmarking Unsupervised Adaptation across Geographies}

\author{Tarun Kalluri
\qquad
Wangdong Xu
\qquad
Manmohan Chandraker \\
UC San Diego \\ 
\texttt{\{sskallur,w6xu,mkchandraker\}@eng.ucsd.edu} \\
\MYhref[violet]{https://tarun005.github.io/GeoNet}{\texttt{https://tarun005.github.io/GeoNet}} \\
% \tk{\textbf{Camera-ready changes in Blue.}}
}
\maketitle

%%%%%%%%%%%%%%%%%%%%%%%%%%%%%%%%%%%%%%%%%%%%%%%%%%%%%%%%%%%%%%%
%%%%%%%%%%%%%%%%%%%%%%%%%%%%%%%%%%%%%%%%%%%%%%%%%%%%%%%%%%%%%%
\begin{abstract}
In recent years, several efforts have been aimed at improving the robustness of vision models to domains and environments unseen during training. An important practical problem pertains to models deployed in a new geography that is under-represented in the training dataset, posing a direct challenge to fair and inclusive computer vision. In this paper, we study the problem of geographic robustness and make three main contributions. 
First, we introduce a large-scale dataset \textbf{\Ours{}} for geographic adaptation containing benchmarks across diverse tasks like scene recognition (\GeoP{}), image classification (\GeoI{}) and universal adaptation (\GeoU{}).
Second, we investigate the nature of distribution shifts typical to the problem of geographic adaptation and hypothesize that the major source of domain shifts arise from significant variations in scene context (context shift), object design (design shift) and label distribution (prior shift) across geographies.
Third, we conduct an extensive evaluation of several state-of-the-art unsupervised domain adaptation algorithms and architectures on \Ours{}, showing that they do not suffice for geographical adaptation, 
and that large-scale pre-training using large vision models also does not lead to geographic robustness. 
Our dataset is publicly available at \MYhref[violet]{https://tarun005.github.io/GeoNet}{\texttt{https://tarun005.github.io/GeoNet}}.
\end{abstract}
%%%%%%%%%%%%%%%%%%%%%%%%%%%%%%%%%%%%%%%%%%%%%%%%%%%%%%%%%%%%%%
%%%%%%%%%%%%%%%%%%%%%%%%%%%%%%%%%%%%%%%%%%%%%%%%%%%%%%%%%%%%%%

%%%%%%%%%%%%%%%%%%%%%%%%%%%%%%%%%%%%%%%%%%%%%%%%%%%%%%%%%%%%%%
%%%%%%%%%%%%%%%%%%%%%%%%%%%%%%%%%%%%%%%%%%%%%%%%%%%%%%%%%%%%%%
    
\section{Introduction}
\label{sec:intro}

\begin{figure}[t]
\begin{center}
    \begin{minipage}[b]{0.42\textwidth}
        \centering
        \includegraphics[width=\textwidth]{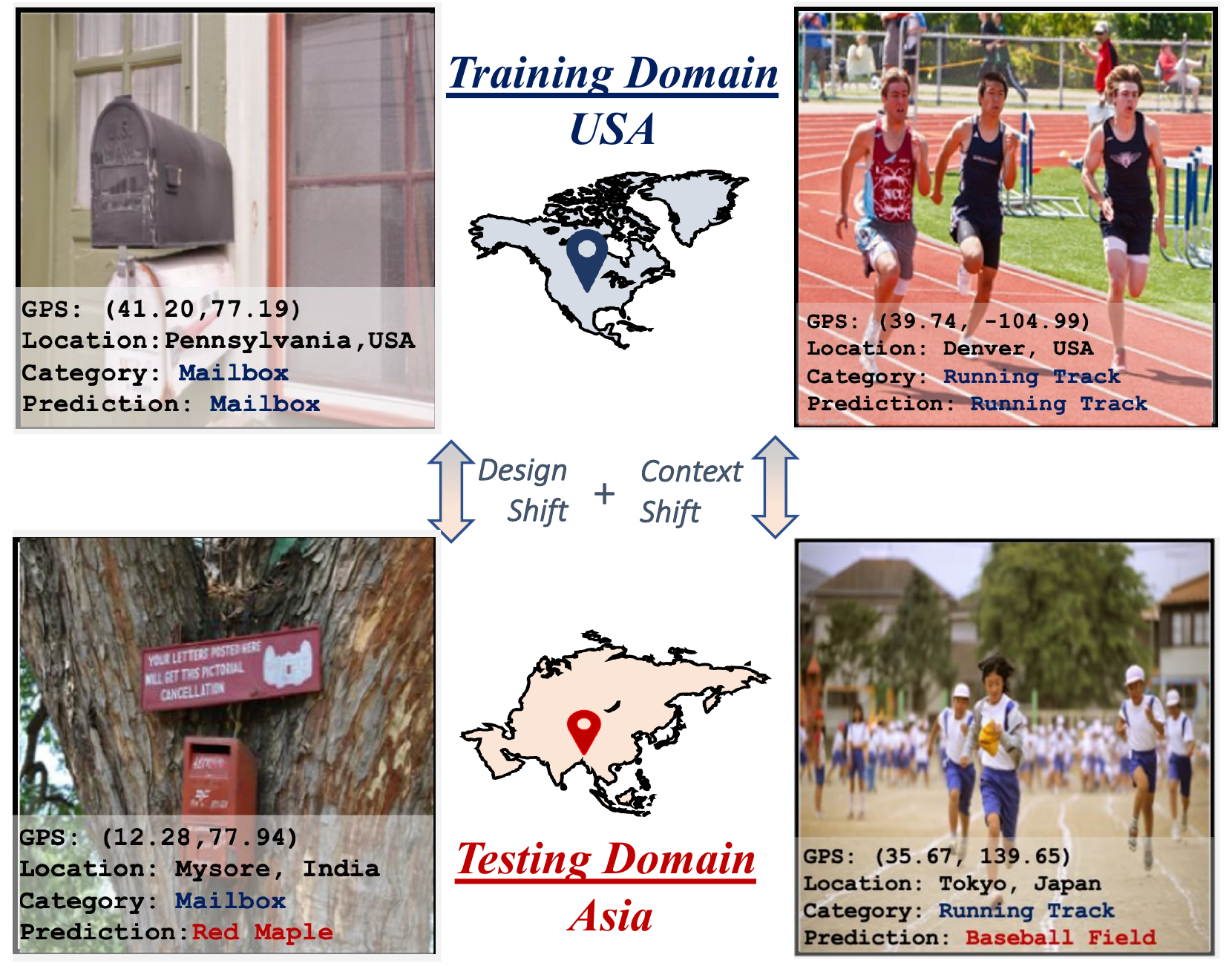}
        % \vspace{-8pt}
        \subcaption{{\bf Geographic bias manifested in proposed \Ours{} dataset} }
        \label{fig:worldmap}
    \end{minipage}
    \vspace{0.15cm}
    \begin{minipage}[b]{0.46\textwidth}
        \centering
        \includegraphics[width=\textwidth]{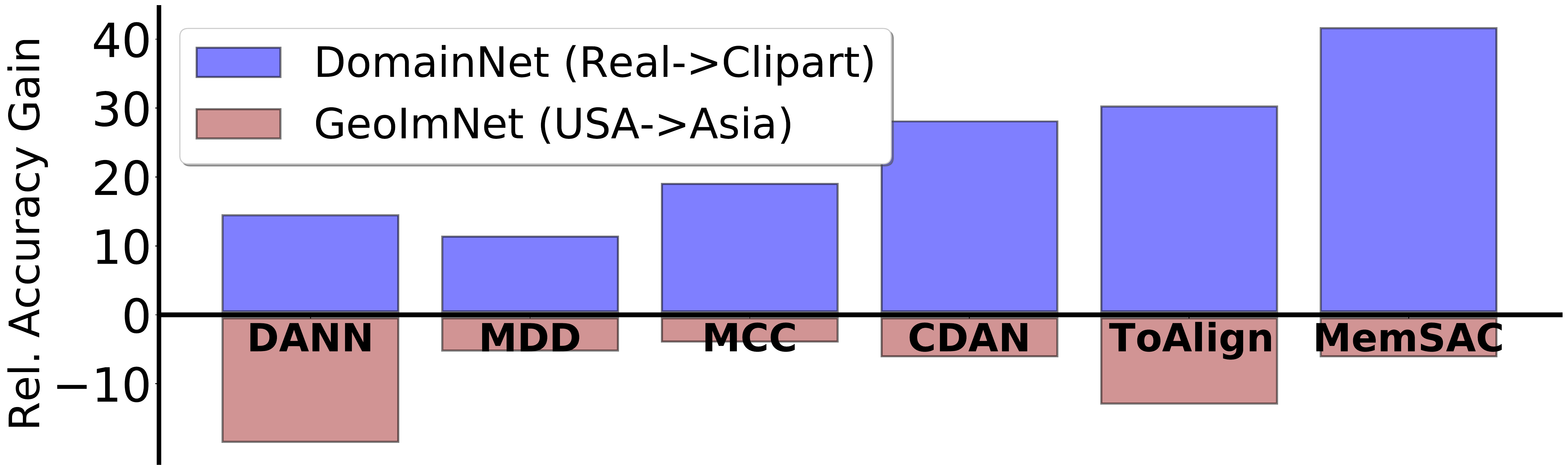}
        \subcaption{{\bf Unsupervised domain adaptation does not suffice on \Ours{}}}
    \label{fig:teaser_adaptation}
    \end{minipage}
    \vspace{0.15cm}
    \begin{minipage}[b]{0.46\textwidth}
        \centering
        \includegraphics[width=\textwidth]{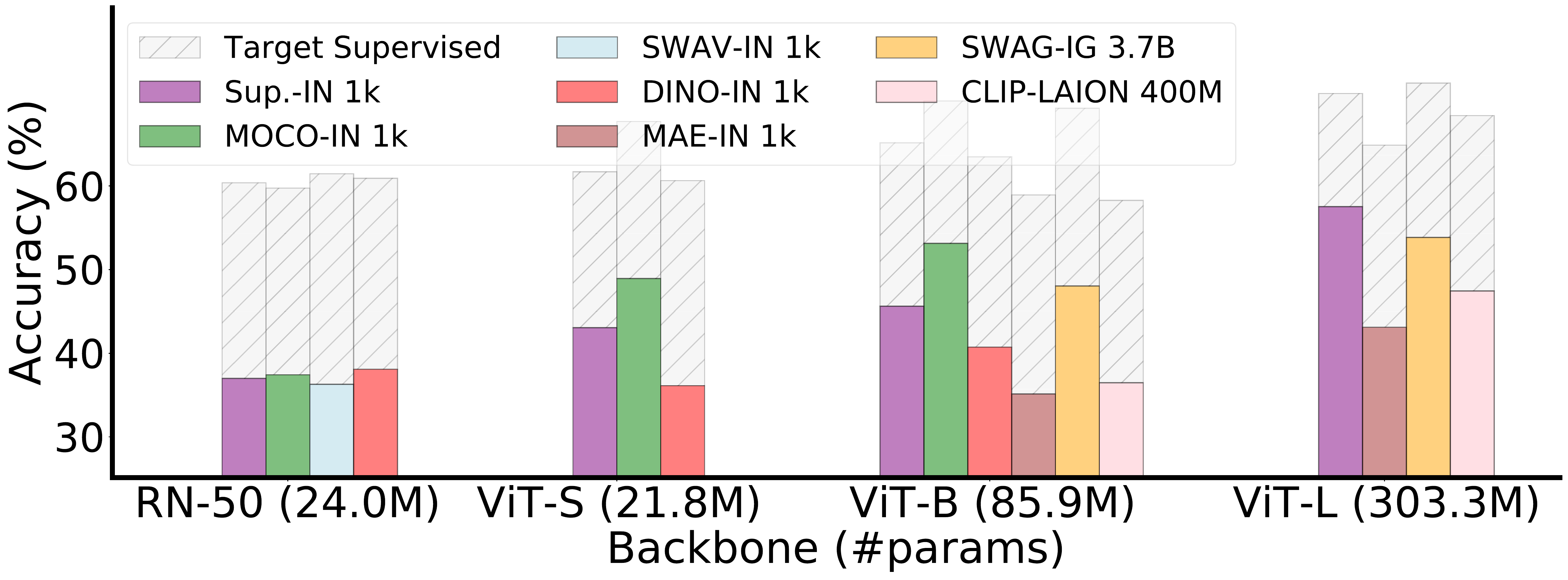}
        \subcaption{{\bf Large vision models exhibit cross-domain drops on \Ours{}}}
    \label{fig:teaser_archs}
    \end{minipage}
    \vspace{-0.1cm}
    \captionsetup{width=0.46\textwidth, font=footnotesize}
    \caption{{\bf Summary of our contributions}. \subref{fig:worldmap}: Training computer vision models on geographically biased datasets suffers from poor generalization to new geographies. We propose a new dataset called \Ours{} to study this problem and take a closer look at the various types of domain shifts induced by geographic variations. \subref{fig:teaser_adaptation} \tk{Prior unsupervised adaptation methods that efficiently handle other variations do not suffice for improving geographic transfer.} \subref{fig:teaser_archs} \tk{We highlight the limitations of modern convolutional and transformer architectures in addressing  geographic bias, exemplified here by  USA$\rightarrow$Asia transfer on \GeoI{}. }}
    \vspace{-3em}
\end{center}
\end{figure}

In recent years, domain adaptation has emerged as an effective technique to alleviate dataset bias \cite{torralba2011unbiased} during training and improve transferability of vision models to sparsely labeled target domains~\cite{long2015learning, long2017deep, CDAN, DANN, saito2017adversarial, saito2018maximum, hoffman2018cycada, xu2019larger, kang2019contrastive, wei2021toalign, kalluri2022memsac}. While being greatly instrumental in driving research forward, methods and benchmark datasets developed for domain adaptation \cite{office31, venkateswara2017deep, peng2017visda, peng2019moment} have been restricted to a narrow set of divergences between domains. 
However, the geographic origin of data remains a significant source of bias, attributable to several factors of variation between train and test data.
 Training on geographically biased datasets may cause a model to learn the idiosyncrasies of their geographies, preventing generalization to novel domains with significantly different geographic and demographic composition. Besides robustness, this may have deep impact towards fair and inclusive computer vision, as most modern benchmark datasets like ImageNet~\cite{ILSVRC15} and COCO~\cite{lin2014microsoft} suffer from a significant US or UK-centric bias in  data~\cite{shankar2017no, DBLP:journals/corr/abs-1906-02659}, with poor representation of images from various other geographies like Asia.

In this paper, we study the problem of geographic adaptation by introducing a new large-scale dataset called \Ours{}, which constitutes three benchmarks -- \GeoP{} for scene classification, \GeoI{} for object recognition and \GeoU{} for universal domain adaptation. These benchmarks contain images from USA and Asia, which are two distinct geographical domains separated by various cultural, economic, demographic and climatic factors. 
We additionally provide rich metadata associated with each image, such as GPS location, captions and hashtags, to facilitate algorithms that leverage multimodal supervision.
\Ours{} captures the multitude of novel challenges posed by varying image and label distributions across geographies. We analyze \Ours{} through new sources of domain shift caused by geographic disparity, namely (i) \emph{context shift}, where the appearance and composition of the background in images changes significantly across geographies, (ii) \emph{design shift}, where the design and make of various objects changes across geographies, and (iii) \textit{prior shift}, caused by different per-category distributions of images in both domains. We illustrate examples of performance drop caused by these factors in \cref{fig:worldmap}, where models trained on images from USA fail to classify common categories such as \textit{running track} and \textit{mailbox} due to context and design shifts, respectively. 

\Ours{} is an order of magnitude larger than previous datasets for geographic adaptation \cite{prabhu2022can,rojasdollar}, allowing the training of modern deep domain adaptation methods. Importantly, it allows comparative analysis of new challenges posed by geographic shifts for algorithms developed on other popular adaptation benchmarks \cite{office31,peng2017visda,venkateswara2017deep,peng2019moment}.
Specifically, we evaluate the performance of several state-of-the-art unsupervised domain adaptation algorithms on \Ours{}, and show their limitations in bridging domain gaps caused by geographic disparities. As illustrated in \cref{fig:teaser_adaptation} for the case of DomainNet~\cite{peng2019moment} vs. \Ours{}, state-of-the-art models on DomainNet often lead to accuracies even worse than a source only baseline on \Ours{}, resulting in negative \textit{relative gain} in accuracy (defined as the gain obtained by an adaptation method over a source-only model as a percentage of gap between a source-only model and the target-supervised upper bound).
Furthermore, we also conduct a study of modern architectures like vision transformers and various pre-training strategies, to conclude that larger models with supervised and self-supervised pre-training offer improvements in accuracy, which however are not sufficient to address the domain gap (\cref{fig:teaser_archs}). This highlights that the new challenges introduced by geographic bias such as context and design shift are relatively under-explored, where our dataset may motivate further research towards this important problem.

In summary, our contribution towards geographic domain adaptation is four-fold:
\begin{tight_itemize}
\item A new large-scale dataset, \Ours{}, with benchmarks for diverse tasks like scene classification and object recognition, with labeled images collected from geographically distant locations across hundreds of categories (\cref{sec:dataset}).
\item Analysis of domain shifts in geographic adaptation, which may be more complex and subtle than style or appearance variations (\cref{subsec:dist_shifts}).
\item Extensive benchmarking of unsupervised adaptation algorithms, highlighting their limitations in addressing geographic shifts (\cref{sec:exp_uda}). 
\item Demonstration that large-scale pretraining and recent advances like vision transformers do not alleviate these geographic disparities (\cref{subsec:pretraining_Exp}).
\end{tight_itemize}

%%%%%%%%%%%%%%%%%%%%%%%%%%%%%%%%%%%%%%%%%%%%%%%%%%%%%%%%%%%%%%
%%%%%%%%%%%%%%%%%%%%%%%%%%%%%%%%%%%%%%%%%%%%%%%%%%%%%%%%%%%%%%

%%%%%%%%%%%%%%%%%%%%%%%%%%%%%%%%%%%%%%%%%%%%%%%%%%%%%%%%%%%%%%
%%%%%%%%%%%%%%%%%%%%%%%%%%%%%%%%%%%%%%%%%%%%%%%%%%%%%%%%%%%%%%

\section{Related Works}
\label{sec:relatedwork}

\vspace{4pt}
\noindent {\bf Domain Adaptation} Unsupervised domain adaptation enables training models on a labeled source domain along with unlabeled samples from a different target domain to improve the target domain accuracy. A large body of prior works aim to minimize some notion of divergence~\cite{ben2006analysis, ben2010theory} between the source and target distributions based on MMD~\cite{tan2020class, long2015learning, long2017deep, sun2016deep} adversarial~\cite{DANN, CDAN, bousmalis2016domain, tzeng2017adversarial, saito2017adversarial, zhang2018collaborative, chen2019progressive, tzeng2015simultaneous}, generative~\cite{sankaranarayanan2018generate, bousmalis2017unsupervised, hoffman2018cycada}, class-level~\cite{pei2018multi, saito2018maximum, luo2019taking, xie2018learning, kumar2018co, gu2020spherical} or instance-level alignment~\cite{wei2021toalign, sharma2021instance, wang2021cross} techniques. Clustering~\cite{deng2019cluster, kalluri2022cluster, kang2019contrastive, park2020joint, kalluri2019universal} and memory-augmentation approaches~\cite{kalluri2022memsac} have also been shown to be effective.
However, most of these works are shown to improve performance using standard datasets such as Office-31~\cite{office31}, visDA~\cite{peng2017visda}, OfficeHome~\cite{venkateswara2017deep} or DomainNet~\cite{peng2019moment}, where the distribution shifts typically arise from unimodal variations in style or appearance between source and target. 
While prior works also study semantic shift~\cite{benmalek2021learning} and sub-population shift~\cite{cai2021theory}, we aim to address a more practical problem of geographic domain adaptation with more complex variations not covered by prior works.

%%%%%%%%%%%%%%%%%%%%%%%%%%%%%%%%%%%%%%%%%%%%%%%%%%%%%%%%
%%%%%%%%%%%%%%%%%%%%%%%%%%%%%%%%%%%%%%%%%%%%%%%%%%%%%%%%
%%%%%%%%%%% Label distribution in source and target %%%%
%%%%%%%%%%%%%%%%%%%%%%%%%%%%%%%%%%%%%%%%%%%%%%%%%%%%%%%%
\begin{figure*}[!t]
     \centering
    \begin{minipage}[b]{0.96\textwidth}
        \centering
        \includegraphics[width=\textwidth]{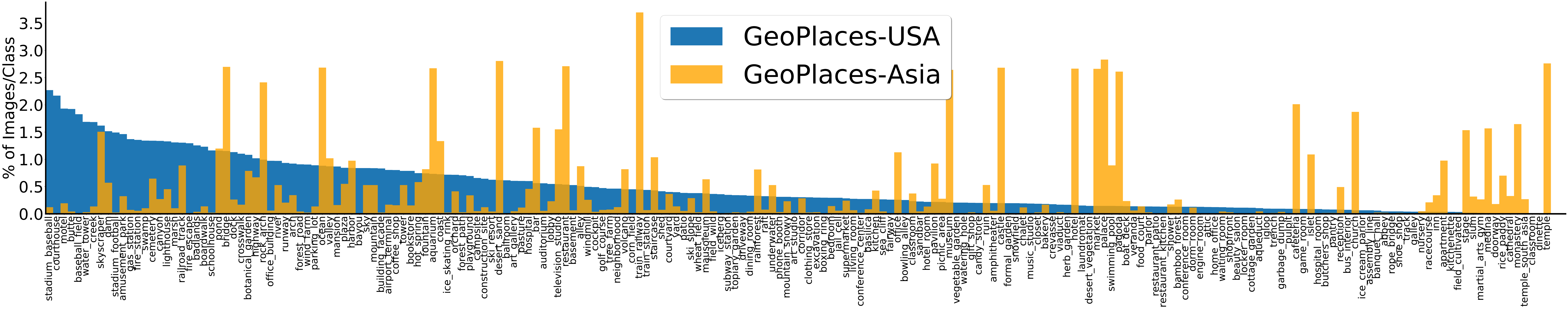}
        \captionsetup{width=0.95\textwidth, font=footnotesize}
        \vspace{-16pt}
        \subcaption{{\bf \GeoP}}
        \label{fig:labeldist_places}
    \end{minipage}
    
    \begin{minipage}[b]{0.96\textwidth}
        \centering
        \includegraphics[width=\textwidth]{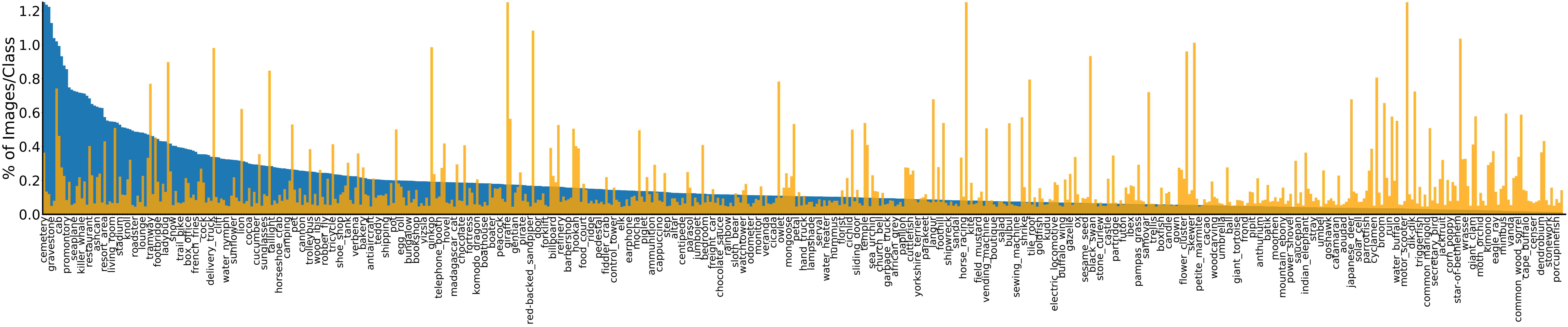}
        \captionsetup{width=0.95\textwidth, font=footnotesize}
        \vspace{-16pt}
        \subcaption{{\bf \GeoI}}
        \label{fig:labeldist_imnet}
     \end{minipage}
     \caption{{\bf Class distribution in \Ours{}} \tk{Percentage of images per class from USA and Asia domains shown for the \GeoP{} benchmark in \subref{fig:labeldist_places} and \GeoI{} benchmark in \subref{fig:labeldist_imnet}. The label distributions are long-tailed in both, and the dominant and tail classes are widely different across geographies in each setting indicating a strong prior shift.} (Best viewed in color, zoom in to see the class names).}
     \label{fig:labeldists}
    \vspace{-12pt}
\end{figure*}
%%%%%%%%%%%%%%%%%%%%%%%%%%%%%%%%%%%%%%%%%%%%%%%%%%%%%%%%
%%%%%%%%%%%%%%%%%%%%%%%%%%%%%%%%%%%%%%%%%%%%%%%%%%%%%%%%

\vspace{4pt}
\noindent {\bf Geographic Robustness} %
Many prior works study biases of CNNs towards 3D poses~\cite{alcorn2019strike, zhao2022ood}, textures~\cite{geirhos2018imagenet}, styles~\cite{hendrycks2021many}, natural variations~\cite{pmlr-v97-recht19a, beyer2020we, taori2020measuring} and adversarial inputs~\cite{hendrycks2021many}, but robustness of computer vision towards shift induced by geography is relatively under-explored.
While algorithms for bridging geographic domain gaps have been proposed in \cite{chen2017no, kalluri2019universal, wang2020train}, they are restricted to road scenes with limited number of classes.
A major hindrance has been the lack of suitable benchmark datasets for geographic adaptation, so several datasets have been recently proposed to address this issue \cite{googlew-49015, DBLP:journals/corr/abs-1906-02659, prabhu2022can, rojasdollar}. Datasets based on dollar street images \cite{rojasdollar} highlight the geographic differences induced by income disparities between various countries, Ego4D~\cite{grauman2022ego4d} contains egocentric videos with actions from various geographies, while researchers in \cite{prabhu2022can} design an adaptation dataset with images from YFCC-100M~\cite{dubey2021adaptive} to analyze geographic shift.
Adding to these efforts, we propose a much larger-scale dataset for geographic adaptation consisting of more diverse categories for place and object classification, across factors of variation beyond income disparities.

%%%%%%%%%%%%%%%%%%%%%%%%%%%%%%%%%%%%%%%%%%%%%%%%%%%%%%%%%%%%%%
%%%%%%%%%%%%%%%%%%%%%%%%%%%%%%%%%%%%%%%%%%%%%%%%%%%%%%%%%%%%%%

%%%%%%%%%%%%%%%%%%%%%%%%%%%%%%%%%%%%%%%%%%%%%%%%%%%%%%%%%%%%%%
%%%%%%%%%%%%%%%%%%%%%%%%%%%%%%%%%%%%%%%%%%%%%%%%%%%%%%%%%%%%%%

\section{Dataset Creation and Analysis}
\label{sec:dataset}

\begin{table}[!t]
  \centering
  \resizebox{0.45\textwidth}{!}{
  % \begin{tabular}{@{} >{\raggedright}p{5.2cm} >{\raggedright}p{2.2cm} >{\raggedright}p{2.2cm} >{\raggedright}p{2cm} p{2cm}}
  \begin{tabular}{cccccc}
    \toprule  \\[-1em]
    % && \multicolumn{2}{c}{Unsupervised Adaptation} && Universal Adaptation \\
    % && \multicolumn{2}{c}{UDA} && UniDA \\
    % \cline{3-4} \cline{6-6}
    & Split &                               \GeoP & \GeoI && \GeoU{}  \\
    % \cline{3-6}
    \midrule
    \multirow{2}{*}{USA}    & Train        & 178110 & 154908 && 100136  \\
                            & Test         & 17234  & 16784 &&  25034  \\
                            % & \#Classes    & 205    & - && 186     \\
    \midrule
    \multirow{2}{*}{Asia}   & Train        & 187426 & 68722 && 33912   \\
                            & Test         & 26923  & 9636 && 8478     \\
                            % & \#Classes    & 205    & - && 185      \\
    \midrule
    \multicolumn{2}{l}{classes-shared}     & 205    & 600 &&  62     \\
    \multicolumn{2}{l}{classes-private}    & -    & - &&  138     \\
    \bottomrule \\
  \end{tabular}
  }
  \vspace{-1em}
  \captionsetup{width=0.45\textwidth}
  \caption{\label{tab:dataset-stats} {\bf Summary of \Ours{}} \tk{Number of images in train and test splits in each of our benchmarks. While \GeoP{} and \GeoI{} are developed for unsupervised adaptation, \GeoU{} is developed for universal domain adaptation across geographies.}}
  \vspace{-12pt}
\end{table}

We present the overall summary of various datasets in our benchmark in \cref{tab:dataset-stats}, including the number of images and categories from each of our settings. In this paper, we broadly consider US and Asia as the two domains, as these two geographies have considerable separation in terms of underlying cultural, environmental and economical factors, while also providing the appropriate level of abstraction and leaving enough data from each domain to perform meaningful analysis.
Although Asia is less homogeneous than USA with greater within-domain variance, our adopted geographical granularity follows from the amount of data we could retrieve from different countries using concepts in \Ours{}, where we observed general paucity in images from many low-resource countries on Flickr.
We also note that the domain shifts caused by geographic disparities are not restricted to these regions, and use images from Africa to show similar observations of domain gaps in the supplementary.

%%%%%%%%%%%%%%%%%%%%%%%%%%%%%%%%%%%%%%%%%%%%%%%%%%%%%%%%
%%%%%%%%%%%%%%%%%%%%%%%%%%%%%%%%%%%%%%%%%%%%%%%%%%%%%%%%
%%%%%%%%%%% Label distribution in source and target %%%%
%%%%%%%%%%%%%%%%%%%%%%%%%%%%%%%%%%%%%%%%%%%%%%%%%%%%%%%%
\begin{figure*}[!t]
    \centering
    \begin{subfigure}[b]{0.48\textwidth}
        \centering
        \includegraphics[width=\textwidth]{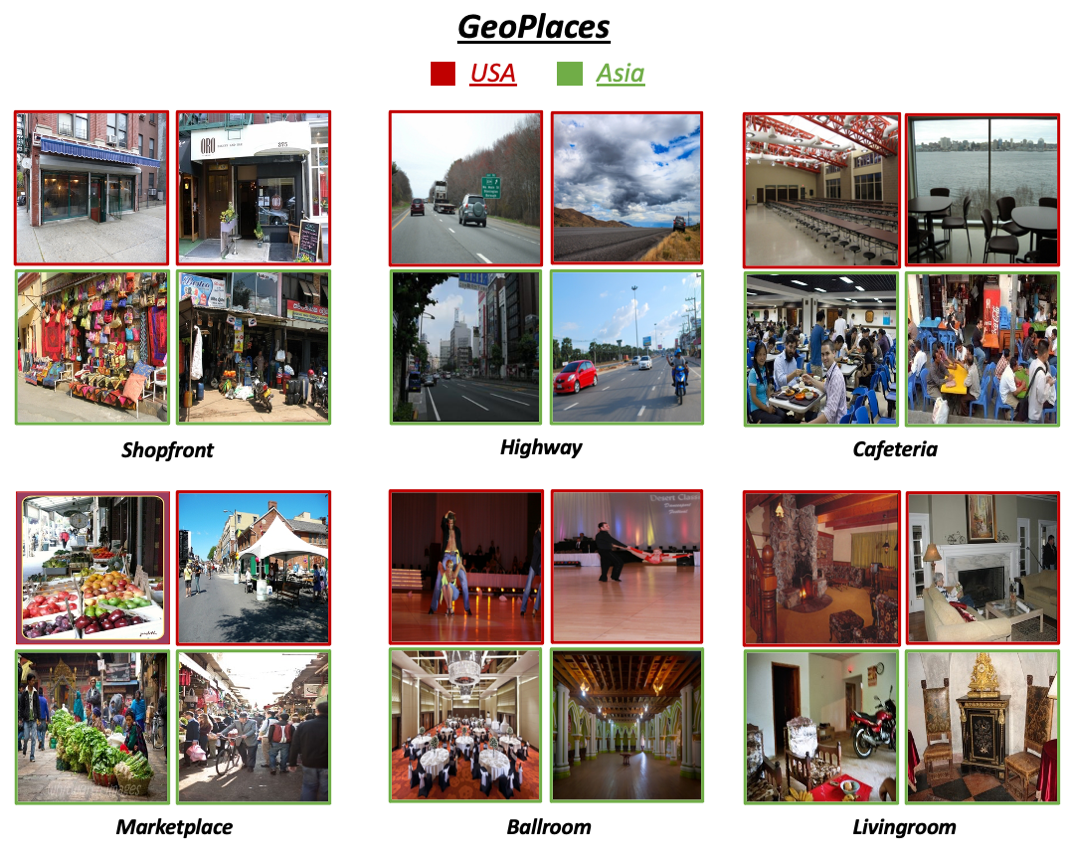}
        \captionsetup{width=0.95\textwidth, font=footnotesize}
        \subcaption{\textbf{\GeoP}}
        \label{fig:contextshift_places}
    \end{subfigure}
    \hfill
    \begin{subfigure}[b]{0.48\textwidth}
        \centering
        \includegraphics[width=\textwidth]{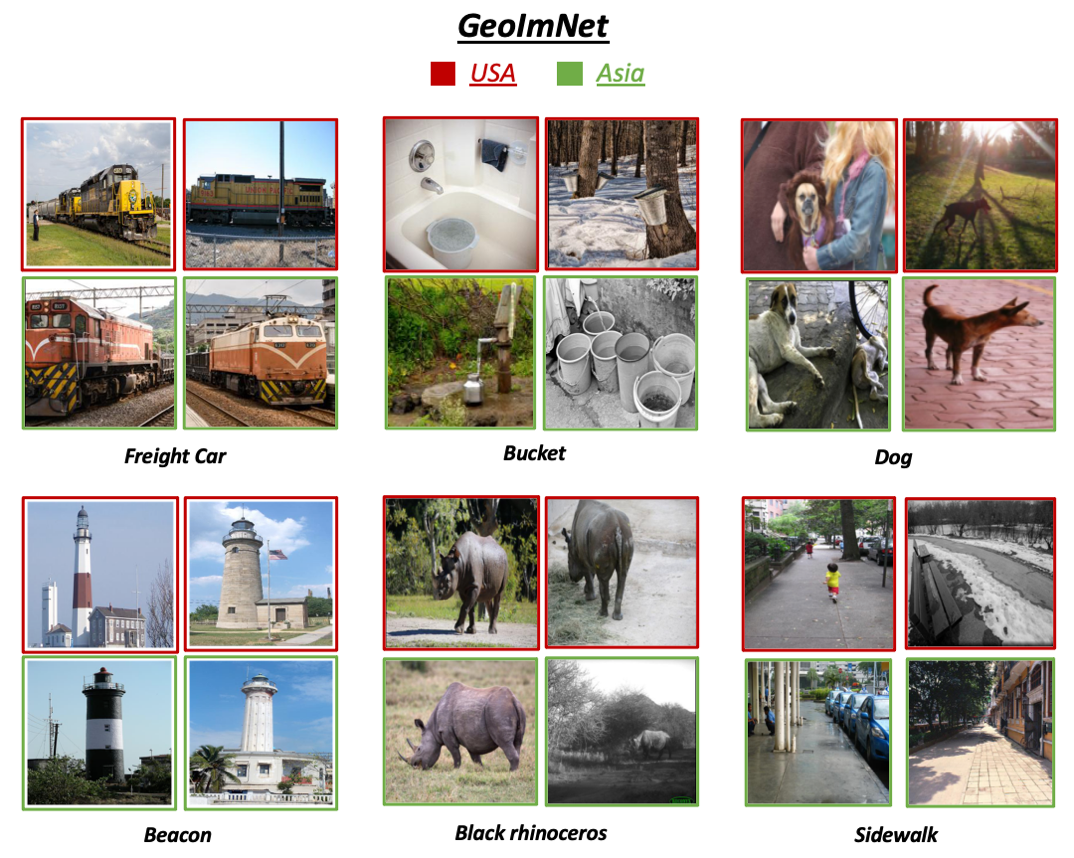}
        \captionsetup{width=0.95\textwidth, font=footnotesize}
        \subcaption{\textbf{\GeoI}}
        \label{fig:contextshift_imnet}
     \end{subfigure}
     % \begin{subfigure}[b]{0.95\textwidth}
     %    \centering
     %    \includegraphics[width=\textwidth]{figures/context_imnet.png}
     %    \captionsetup{width=0.95\textwidth, font=footnotesize}
     %    \subcaption{Design Shift}
     %    \label{fig:designshift_imnet}
     % \end{subfigure}
    \vspace{-5pt}
     \caption{{\bf Context Shift in \Ours{}} A few examples showing the nature of context shifts across categories from \GeoP{} benchmark in \subref{fig:contextshift_places}, and \GeoI{} benchmark in \subref{fig:contextshift_places}, arising due to a variety of differences between geographical disparity. 
     For example, outdoor scenes (shopfront, marketplace) reflect the demographies across geographies, indoor-scenes (living rooms, cafeteria) reflect cultural and economic variations and wildlife images reflect the habitat and climatic variations. }
     \label{fig:contextshift_illus}
    \vspace{-12pt}
\end{figure*}
%%%%%%%%%%%%%%%%%%%%%%%%%%%%%%%%%%%%%%%%%%%%%%%%%%%%%%%%
%%%%%%%%%%%%%%%%%%%%%%%%%%%%%%%%%%%%%%%%%%%%%%%%%%%%%%%%
\begin{figure*}[!t]
    \centering
    \includegraphics[width=\textwidth]{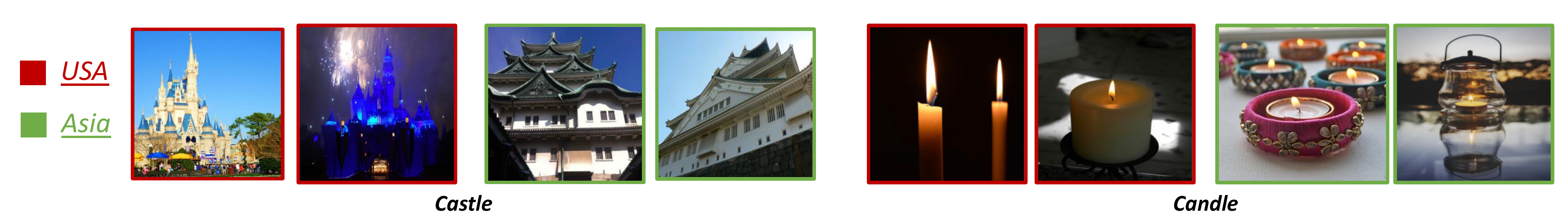}
    \captionsetup{width=\textwidth, font=footnotesize}
    \vspace{-16pt}
     \caption{{\bf Design Shift in \Ours{}} We show examples illustrating the design shifts for the cases of \textit{castle} from \GeoP{} and \textit{candle} from \GeoI{}. Note that differences in designs of castles as well as the variety of objects like candles found across geographies lead to design shifts between the domains.
     }
     \label{fig:designshift_illus}
    \vspace{-12pt}
\end{figure*}
%%%%%%%%%%%%%%%%%%%%%%%%%%%%%%%%%%%%%%%%%%%%%%%%%%%%%%%%
%%%%%%%%%%%%%%%%%%%%%%%%%%%%%%%%%%%%%%%%%%%%%%%%%%%%%%%%

\subsection{\GeoP{}} 
\label{subsec:geoPlaces}

\noindent We propose \GeoP{} to study geographic adaptation in scene classification, which involves predicting the semantic category of the place or location present in the image~\cite{zhou2017places}. In contrast to object classification, it is necessary to accurately identify and understand various interactions and relationships between the objects and people in the scene to predict the appropriate scene category. 
In spite of rapid progress in datasets~\cite{xiao2010sun,zhou2017places} and methods~\cite{NEURIPS2021_27d52bcb} for this task, robustness of scene classification networks to unseen domains in general, and across geographies in particular, has received little attention, for which we propose a suitable benchmark.

\vspace{2pt}
\noindent \textbf{Selecting Concepts and Images} We use the 205 scene categories from Places-205~\cite{zhou2017places} to build \GeoP{}, as these semantic categories cover a wide range of real world scenes commonly encountered in most geographies. 
We build our \GeoP{} benchmark from the labeled Places-205 dataset~\cite{zhou2014learning}. 
We first collect the unique Flickr identifier (\texttt{Flickr-id}) associated with each image in the Places-205 dataset, and
then use the publicly available Flickr API\footnote{\href{https://www.Flickr.com/services/api/explore/Flickr.photos.geo.getLocation}{Flickr.com/services/api/explore/Flickr.photos.geo.getLocation}} to extract the GPS location of the image.
Since only a fraction of images belong to Flickr and a further smaller fraction contain valid geotags, we end up with around 400k images from 205 classes with associated geographical information. 
Of these, 190k images are from the US domain, and we use 178k of them for training and 17k for testing. In Asia domain however, we obtain only 27k images. To match the scale of images from both domains, we perform an additional step and manually collect more images as explained next.

\vspace{4pt}
\noindent \textbf{Additional Data} Due to the inherent US-centric bias of photo-sharing websites like Flickr, a major portion of images are US-based. In order to collect more images from the Asia domain, we directly scrape images from Flickr using the 205 category names from Places-205 as the \textit{seed concepts}. 
As many Asian users often post descriptions and tags for pictures in languages other than English, we use translations of these seed concepts in English to 6 Asian languages, namely \{Hindi, Korean, Japanese, Chinese, Russian, Hebrew\}, and use these along with the original concepts, as the augmented or \textit{expanded concepts}. Then, we search Flickr for images which match the criterion that (i) they are geotagged in Asia, and (ii) the tags associated with the image match with exactly one of the categories in the expanded concept list (which we assign as the label). We collect around 190k images this way, and use this as the training set. Since images collected from web tend to be nosier than human labeled ones, we use the manually labeled 27k images from Places-205 as the test set for Asia domain to ensure robust benchmarking. 

\subsection{\GeoI{}} 
\label{subsec:geoImages}

We propose the \GeoI{} benchmark to investigate the domain shift due to geographical disparities on object classification.
Different from existing object-level datasets for domain adaptation~\cite{peng2017visda, peng2019moment, office31, venkateswara2017deep}, \GeoI{} provides domain shifts induced by geographic disparities.

\vspace{2pt}
\noindent \textbf{Dataset curation} We collect images in the \GeoI{} benchmark from the WebVision dataset~\cite{li2017webvision}, which itself is scraped from Flickr using queries generated from 5000 concepts in the Imagenet-5k dataset~\cite{deng2009imagenet}. 
We then follow the same pipeline as explained above for \GeoP{} benchmark, and identify the GPS coordinates of each images using its \texttt{Flickr-id}.

\vspace{2pt}
\noindent \textbf{Concept Selection} Although the original dataset contains 5000 classes, many of these classes are indigenous to a particular geography. For example, \textit{Bengal Tiger}s are found in Indian subcontinent, and \textit{Bald Eagle} is a North-American bird. Since unsupervised domain adaptation typically demands matching label spaces across source and target, we select 600 categories out of the original 5000 with at least 20 images in each domain from each category. We then assign roughly 15\% of images from each domain into the test set and use the remaining as the training images. 

\vspace{2pt}
\noindent \textbf{Dataset filtering} WebVision is \textit{webly supervised}~\cite{chen2015webly}, which does not guarantee object-centric images or clean labels. 
Therefore, we remove all the images from the dataset which have more than one tag that match our selected concepts (the 600 chosen categories) to handle multi-labeled images. Furthermore, we manually quality-check all the test images and remove all the images with noisy labels. Finally, we perform de-duplication to remove images from the training set which are very similar to those in the test set. 
More insights into each step of our data collection and filtering process is provided in the \supp{}.
\review{The final label distribution for both US and Asia domains in both our benchmarks is shown in \cref{fig:labeldists}. }

\subsection{\GeoU{}}
\label{subsec:unida}

Universal Domain Adaptation (UniDA) \cite{you2019universal} facilitates domain adaptation between source and target domains that have few {private classes}, in addition to {shared classes} which are common to both. 
While this is a realistic problem, prior works~\cite{you2019universal, saito2021ovanet, saito2020universal, kundu2022subsidiary} use benchmarks created from existing UDA datasets for evaluation. However, our proposed geographical adaptation setting gives us an unique opportunity to design benchmarks for UniDA such that the private categories from the source and the target are a natural reflection of the presence or absence of these categories in the respective geographical domains. 
In order to select the shared and private categories for our Geo-UniDA benchmark, we first start with the 1000 categories in the original Imagenet-1k dataset~\cite{ILSVRC15}, and select top 200 categories each in the USA and Asia domains that have the most number of images from the WebVision dataset. Out of these, we use the 62 common classes as the shared categories, and the remaining 138 as the private classes in each domain.
  
\subsection{Analysis of Distribution Shifts}
\label{subsec:dist_shifts}

We denote the source dataset using $D_s {=} \{X_s,Y_s\}$, and 
assume that $X_s {\sim} P_s(x)$ and $(X_s,Y_s) {\sim} P_s(x,y)$ where $P_s(x)$ and $P_s(x,y)$ are the image marginal and image-label joint distribution respectively.
% is the distribution of the source image space, and $P_s(x,y)$ is the joint distribution of source images and labels. 
Target dataset $D_t = \{X_s,Y_s\}$ and target distributions $P_t(x)$ and $P_t(x,y)$ are defined similarly, and the domain discrepency assumption states that $P_s(x,y) \neq P_t(x,y)$.
In order to formulate domain shift across geographies, we define $f_x$ as the part of image referring to the foreground objects (corresponds to the salient objects in a scene) and $b_x$ to be the rest of the image corresponding to the background regions (corresponding to the surrounding regions or context). 
For example, for the task of classifying \textit{living room} in \cref{fig:contextshift_places} from \GeoP{}, common objects like sofa and table are foreground, while floor, roof and walls are backgrounds. 
% For classifying \textit{bucket} in \cref{fig:contextshift_imnet} from \GeoI{}, the bucket is the foreground while all other regions in the image are backgrounds. 
We make a simplifying assumption that an image is completely explainable using its foreground and background and replace the class-conditional distribution of the images $P(x|y)$ with the joint class-conditional $P(b_x,f_x|y)$. Further, we also assume that given a class label, the background is conditionally independent of the foreground. Then,
\begin{eqnarray}
    P(x,y) &=&  P(x|y) \cdot P(y) \nonumber \\
           &=&  P(b_x,f_x|y) \cdot P(y) \nonumber \\
           &=&  P(b_x|y) \cdot P(f_x|b_x,y) \cdot P(y) \nonumber \\
    \implies P(x,y) &=&  \underbrace{P(b_x|y)}_\text{context} \cdot \underbrace{P(f_x|y)}_\text{design} \cdot \underbrace{P(y)}_\text{prior}
    \label{eq:dshift}
\end{eqnarray}

We define the class-conditional background distribution $P(b_x|y)$ as context, class-conditional object distribution $P(f_x|y)$ as design and the label distribution $P(y)$ as prior. 
% Therefore, the domain shifts between source and target $P_s(x,y)$ and $P_t(x,y)$ can be induced by any one of the factors in \cref{eq:dshift}, and 
Note that standard covariate shift assumption~\cite{ben2010theory} assumes uniform domain discrepency across all the images ($P_s(x) {\neq} P_t(x)$), which does not hold for geographic adaptation due to the diverse source of variations. 
We analyze each of these from a geographic adaptation perspective next.
% We next analyze each of these from a geographic adaptation perspective as reflected in \Ours{}. 

\vspace{2pt}
\noindent {\bf Context Shift} We define context shift to be the changes in the context around an object or scene given by $P_s(b_x|y) \neq P_t(b_x|y)$. 
% natural changes in the scene compositions between diverse geographies leads to notable differences in the context around an object or scene, which we refer to as the {context shift} between the domains, 
Deep learning models are generally sensitive to object contexts and backgrounds, and learn spurious correlations that impede their ability to recognize objects and scenes in novel contexts~\cite{choi2012context, choi2019can, singh2020don, rosenfeld2018elephant}. 
% Even when the context is important for classification in tasks like scene recognition, varying scene constituents across geographies hurts generalization ability. 
In geographic adaptation, context shift can be caused by differences in cultural or economic factors across geographies, and few examples illustrating context shift from \GeoP{} and \GeoI{} are shown in \cref{fig:contextshift_illus}. 
While prior works already introduce context shift for domain adaptation~\cite{prabhu2022can}, a key difference lies in their modeling assumption that the context is irrelevant while training, while in our case context might play a key role in improving scene classification on \GeoP{}. 

\vspace{2pt}
\noindent {\bf Design Shift} We define ``{design}'' shift as the change in object structure, shape and appearance, where the foreground objects belonging to the same semantic category look different across geographies, given by $P_s(f_x|y) \neq P_t(f_x|y)$. Few examples are shown in \cref{fig:designshift_illus}, where categories like \textit{castle} from \GeoP{} and \textit{candle} from \GeoI{} datasets look widely different due to high intra-class variance, although they belong to the same semantic category. 
It is important to note that context and design shifts might also occur within a domain or within a geography. However, it is easier to account for intra-domain variations on labeled source datasets than ensuring robustness to new and unlabeled geographies.

\vspace{2pt}
\noindent {\bf Prior Shift} The label distributions across the domains in our benchmarks widely differ due to natural prominence or rarity of the classes according to the geography, as shown in \cref{fig:labeldists}, where the head classes of one domain might be tail classes in another. This leads to a prior shift where $P_s(y) \neq P_t(y)$. For example, categories like \textit{railway station, outdoor markets, monasteries} are common in Asia while \textit{baseball stadiums} are more common in USA. Prior works examining prior shift or label shift across domains~\cite{azizzadenesheli2019regularized, garg2020unified, zhang2013domain, lipton2018detecting, alexandari2020maximum} generally assume that the class conditionals remain the same, which is not true in the case of geographic adaptation due to context and design shifts as illustrated above. 

%%%%%%%%%%%%%%%%%%%%%%%%%%%%%%%%%%%%%%%%%%%%%%%%%%%%%%%%%%%%%%
%%%%%%%%%%%%%%%%%%%%%%%%%%%%%%%%%%%%%%%%%%%%%%%%%%%%%%%%%%%%%%

\section{Experiments}
\label{sec:expmnts}

\begin{table}[!t]
  \centering
  \resizebox{0.46\textwidth}{!}{
  % \begin{tabular}{@{} >{\raggedright}p{5.2cm} >{\raggedright}p{2.2cm} >{\raggedright}p{2.2cm} >{\raggedright}p{2cm} p{2cm}}
  \begin{tabular}{lccccccccc}
    & \multicolumn{3}{c}{\textbf{\underline{\GeoP}}}  \\
    Train $\downarrow$ / Test $\rightarrow$ & USA & Asia & Drop (\%) \\
    \cline{2-4}
    USA & 56.35/85.15 & 36.27/63.27 & \textcolor{brown}{-20.08/-21.88} \\
    Asia & 21.03/44.81 & 49.63/78.45 & \textcolor{brown}{-28.60/-33.64} \\
    % \textcolor{gray}{USA+Asia} & \textcolor{gray}{yy.yy/yy.yy} & xx.xx/xx.xx &  \\
    \midrule 
    & \multicolumn{3}{c}{\textbf{\underline{\GeoI}}} \\
    Train $\downarrow$ / Test $\rightarrow$ & USA & Asia & Drop (\%) \\
    \cline{2-4}
    USA & 56.35/77.95 & 36.98/63.42 & \textcolor{brown}{-19.37/-14.53} \\
    Asia & 40.43/64.60 & 60.37/80.22 & \textcolor{brown}{-19.94/-15.62} \\
    % \textcolor{gray}{USA+Asia} & \textcolor{gray}{yy.yy/yy.yy} & \textcolor{gray}{xx.xx/xx.xx} &  \\
    \bottomrule
  \end{tabular}
  }
  \vspace{-1em}
  \captionsetup{width=0.45\textwidth}
  \caption{\label{tab:source-transfer} \tk{Top-1/Top-5 accuracies of Resnet-50 models across geographically different train and test domains. Note the significant drop in accuracies caused by the geographical domain shifts in each setting.}}
  \vspace{-.5em}
\end{table}

\begin{table}[!t]
  \centering
  \resizebox{0.46\textwidth}{!}{
  % \begin{tabular}{@{} >{\raggedright}p{5.2cm} >{\raggedright}p{2.2cm} >{\raggedright}p{2.2cm} >{\raggedright}p{2cm} p{2cm}}
  \begin{tabular}{lccccccccc}
    & \multicolumn{3}{c}{Original} && \multicolumn{3}{c}{Balanced} \\
    \cline{2-4} \cline{6-8}
    & USA & Asia & $\Delta$ && USA & Asia & $\Delta$ \\
    \GeoP{} & 56.35 & 36.27 & \textcolor{brown}{20.08\%} && 55.52 & 42.6 &  \textcolor{brown}{12.92\%}  \\
    \GeoI{} & 56.35 & 36.98 &  \textcolor{brown}{19.37\%} && 52.72 & 37.3 &  \textcolor{brown}{15.42\%} \\
    \bottomrule
  \end{tabular}
  }
  \vspace{-1em}
  \captionsetup{width=0.45\textwidth}
  \caption{\label{tab:balanced-acc} \tk{USA ${\rightarrow}$ Asia comparison between \Ours{} and its label-balanced version. Non-trivial gaps between the geographies still exist even after accounting for prior shift between the domains.} }
  \vspace{-18pt}
\end{table}

\subsection{Domain Shifts in Proposed Datasets}
\label{sec:source_transfer}
We illustrate the severity of domain differences across geographies using the drop in accuracy caused by cross-geography transfer in \cref{tab:source-transfer}. Specifically, we train a Resnet-50~\cite{he2016deep} model using images only from one domain, and compute the accuracies on both within-domain and cross-domain test sets. Since a lot of categories in \Ours{} are close (example, \textit{train station} vs. \textit{subway station}), we use both top-1 and top-5 accuracies to report the performance.
We observe a significant drop in accuracy caused by direct transfer of models across domains which can be attributed to the geographic bias in the training data. For example, a model trained on \GeoP{} benchmark on US images gives 56.35\% Top-1 accuracy on US images, but only 36.27\% on images from Asia with a notable drop of 20\%. 
On the \GeoI{} benchmark, within-domain testing on images collected from USA gives 56.35\% top-1 accuracy while cross-domain testing on Asia images gives only 36.98\% with a drop of 19.37\%. The 36.98\% accuracy is also much inferior to the supervised accuracy on the Asia domain (60.37\%) which can be considered as the target upper bound.

\begin{figure}[t]
\begin{center}
        \centering
        \includegraphics[width=0.45\textwidth]{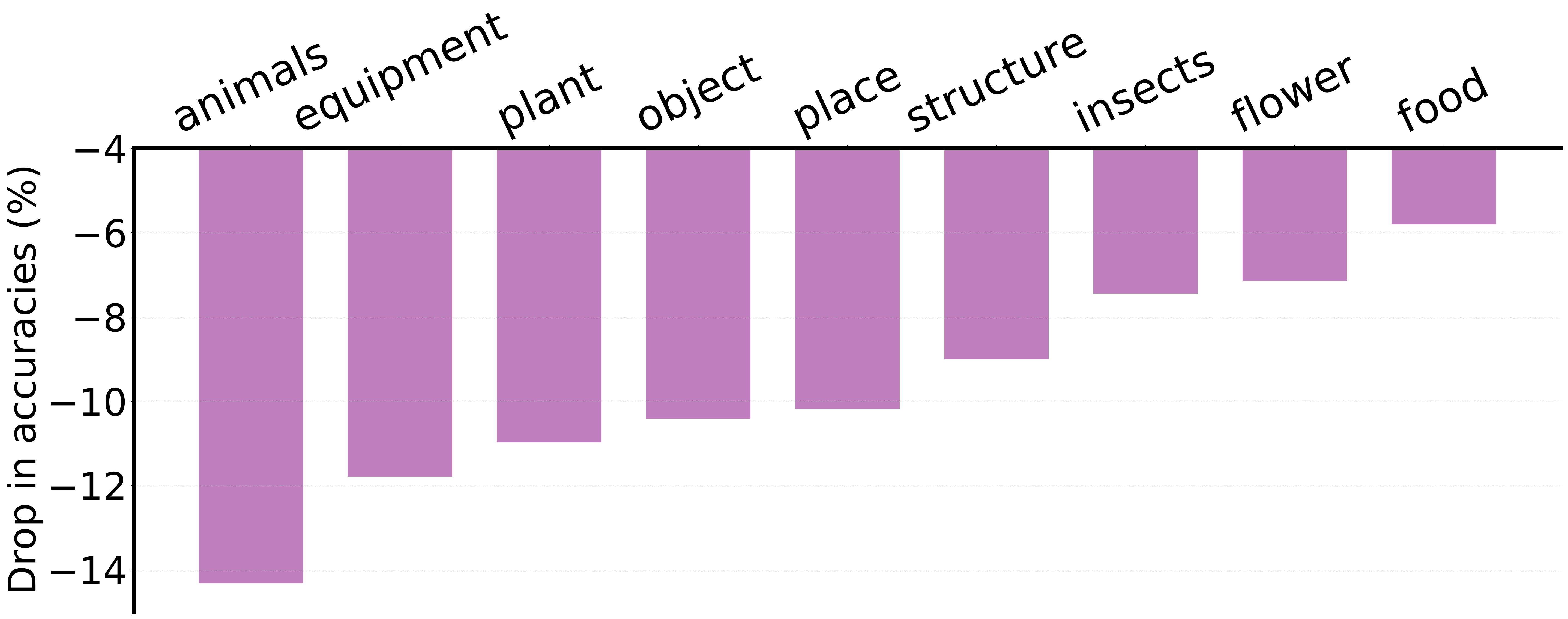}
        \caption{\label{fig:imnet-groupdrops}{Drop in accuracies for each meta-category in \GeoI{}. Groups that showcase context and design shifts suffer a larger drop in accuracy.}
        }
        \vspace{-16pt}
\end{center}
\end{figure}
\begin{figure}[t]
\begin{center}
        \centering
        \includegraphics[width=\columnwidth]{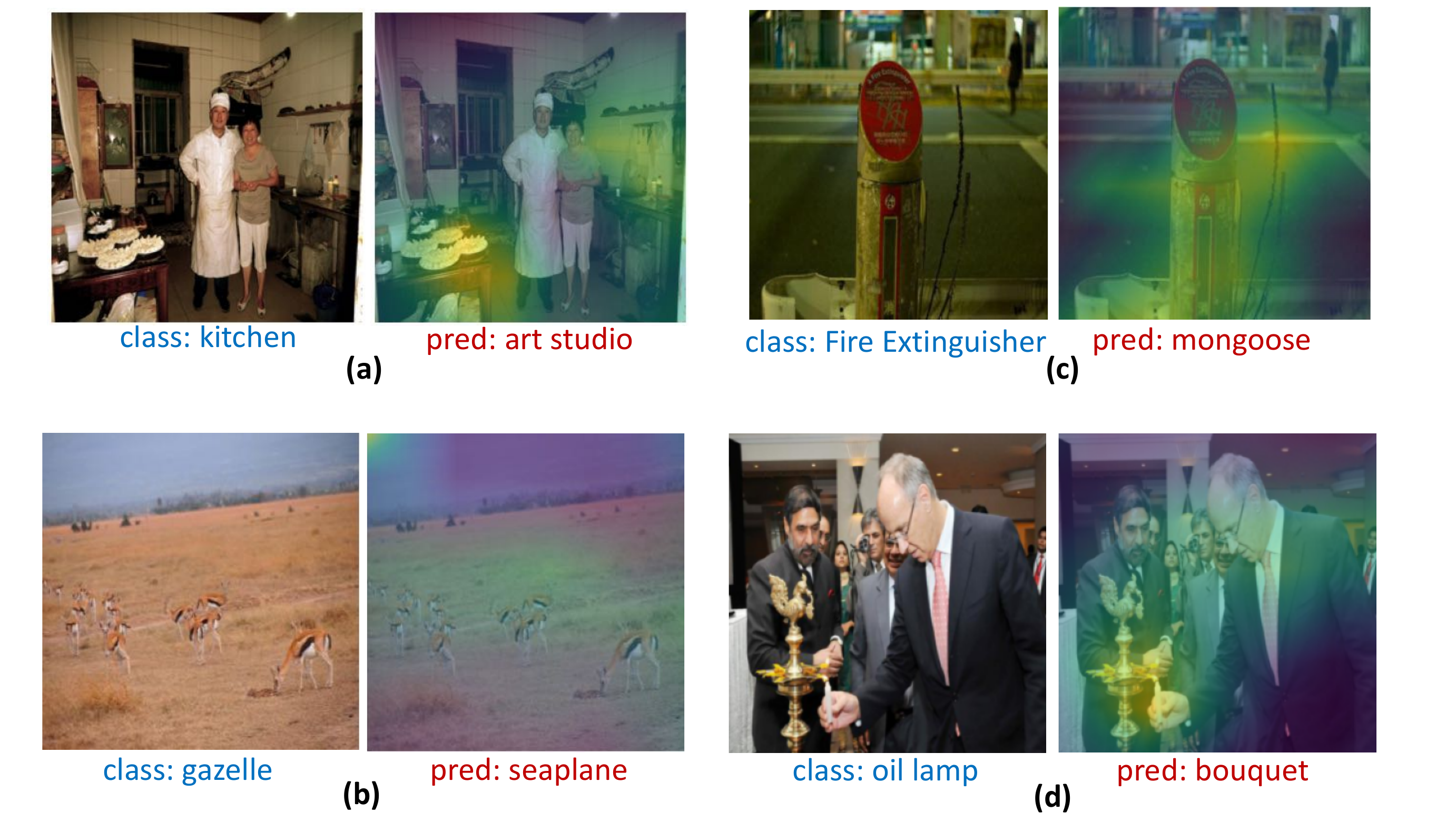}
        \vspace{-2em}
        \caption{\label{fig:gradcams}{GradCAM visualization of predictions of a USA-trained model on Asia images show that prominent context and design shifts across geography hurts accuracy. (a) is from \GeoP{}, (b,c,d) are from \GeoI{}.}
        }
        \vspace{-2em}
\end{center}
\end{figure}

\noindent {\bf Meta-category wise error analysis for \GeoI{}} We relate the drop in performances across geographies to the proposed notions of domain discrepency in geographic adaptation like context and domain shifts in \cref{fig:imnet-groupdrops}. 
Specifically, since the concepts in \GeoI{} are sourced from ILSVRC, we leverage the wordnet hierarchy to group our 600 classes into 9 meta-labels. 
We then average the accuracy within each meta-class from USA$\rightarrow$Asia domain transfer, and plot the difference in accuracy across domains per meta-label in \cref{fig:imnet-groupdrops}. We note that categories in the meta-label ``{animals}'' have minimum design-shift across domains, but suffer from context shift due to shifts in weather and habitats across geographies leading to significant drop in accuracy. On the other hand, many categories in ``{equipment}'' and ``{object}''(like \textit{candle, broom, sewing machine}) have prominent design shifts (\cref{fig:designshift_illus}) leading to notable performance drop. 
Finally, categories in ``food'' (like \textit{bottled water, ice-cream}) have minimum change in both design and context and hence suffer the least fall in accuracy across domains. 

\begin{table*}[!t]
  \centering
  % \rowcolors{2}{gray!25}{white}
  \resizebox{0.75\textwidth}{!}{
  % \begin{tabular}{@{} >{\raggedright}p{5.2cm} >{\raggedright}p{2.2cm} >{\raggedright}p{2.2cm} >{\raggedright}p{2cm} p{2cm}}
  \begin{tabular}{lccccccccccc}
  \toprule
    Method & \multicolumn{4}{c}{\underline{\GeoP}} && \multicolumn{4}{c}{\underline{\GeoI}}  \\
    % \cline{2-12}
             & \multicolumn{2}{c}{USA $\rightarrow$ Asia} & \multicolumn{2}{c}{Asia $\rightarrow$ USA} && \multicolumn{2}{c}{USA $\rightarrow$ Asia} & \multicolumn{2}{c}{Asia $\rightarrow$ USA} \\
             \midrule
                                    & Top-1 & Top-5 & Top-1 & Top-5 && Top-1 & Top-5 & Top-1 & Top-5 \\
                                    \cline{2-5} \cline{7-10}
    % \midrule
     Source Only                    & \textbf{36.27} & \textbf{63.27} & \textbf{21.03} & \textbf{44.81} && \textbf{36.98} & \textbf{63.43} & \textbf{40.43} & \textbf{64.6}\\
     DANN~\cite{DANN}               & 29.58 & 55.23 & 16.59 & 35.32 && 32.88 & 57.77 & 38.42 & 62.90 \\
     CDAN~\cite{CDAN}               & 30.48 & 55.94 & 17.01 & 36.26 && 35.94 & 60.21 & 39.88 & 63.74 \\
     MCC~\cite{jin2020minimum}      & 30.09 & 55.85 & 17.17 & \underline{36.85} && 35.71 & 60.48 & 39.86 & 64.00 \\
     SAFN~\cite{xu2019larger}       & 32.50 & 57.93 & 14.34 & 35.68 && 32.40 & 58.43 & 36.26 & 61.58 \\
     MDD~\cite{zhang2019bridging}   & 34.18 & 59.10 & \underline{17.81} & 36.44 && 36.26 & 62.13 & 40.15 & 63.91 \\
     MCD~\cite{saito2018maximum}    & 33.49 & 59.41 & 16.57 & 34.74 && 25.60 & 48.45 & 36.69 & 60.68 \\
     ToAlign~\cite{wei2021toalign}  & 29.86 & 56.16 & 16.32 & 33.58 && 32.13 & 58.64 & 37.98 & 63.17 \\
     MemSAC~\cite{kalluri2022memsac}& \underline{34.68} & \underline{60.52} & 15.75 & 32.83 && \underline{36.71} & \underline{63.16} & \underline{40.34} & \underline{64.40} \\ 
     \midrule
     \textcolor{gray}{Tgt. Supervised} & \textcolor{gray}{49.63} & \textcolor{gray}{78.45} &\textcolor{gray}{56.35} & \textcolor{gray}{85.15} && \textcolor{gray}{60.37} & \textcolor{gray}{80.22} & \textcolor{gray}{56.35} & \textcolor{gray}{77.95} \\
    \bottomrule
  \end{tabular}
  }
  \vspace{-6pt}
  \captionsetup{width=\textwidth}
  \caption{\label{tab:uda-benchmarking} {\bf UDA on \Ours{}} \tk{Top-1 and Top-5 accuracies of various unsupervised adaptation methods on \Ours{}. Most of the methods fail to sufficiently handle cross-geography transfer on both \GeoP{} and \GeoI{} benchmarks and often give lower accuracies even compared to a baseline model trained only using source data calling attention to the need for novel methods that can handle domain shifts beyond style and appearance. } }
  \vspace{-8pt}
\end{table*}

\begin{table}[!t]
  \centering
  \resizebox{0.46\textwidth}{!}{
  % \begin{tabular}{@{} >{\raggedright}p{5.2cm} >{\raggedright}p{2.2cm} >{\raggedright}p{2.2cm} >{\raggedright}p{2cm} p{2cm}}
  \begin{tabular}{l cc c|cc}
    Method & closed-set & open-set & H-Score && Target Sup.  \\
    \midrule
    UniDA~\cite{you2019universal} & 27.64 & 43.93 & 33.93 && \multirow{3}{*}{\textcolor{gray}{70.70\%}}  \\
    DANCE~\cite{saito2020dance} & 38.54 & 78.73 & 51.75 &&  \\
    OVANet~\cite{saito2021ovanet} & 36.54 & 66.89 & 47.26 &&  \\
    \bottomrule
  \end{tabular}
  }
  \vspace{-1em}
  \captionsetup{width=0.45\textwidth}
  \caption{\label{tab:uniDA-benchmarking} {\bf Universal domain adaptation methods on \GeoU{}}. \textit{closed-set} and \textit{open-set} refer to the closed set and open set accuracies, and \textit{H-Score} is the harmonic-mean of the two. Note the significant gap that still exists with target supervised accuracy on closed-set labels with the best adaptation method DANCE~\cite{saito2020dance}.}
  \vspace{-1em}
\end{table}

\noindent {\bf GradCAM visualization of the failure cases} We present few examples in \cref{fig:gradcams} of predictions made on Asia test images by a model trained on USA, along with their GradCAM visualizations. As shown, when the model focuses on the context and background, it fails to generalize to new scenes from target geographies with notable shifts in context (\textit{kitchen} classified as \textit{art studio}). Even in cases when the model accurately focuses on the foreground object, it sometimes leads to incorrect predictions due to design shifts between geographies, where \textit{oil lamp} is accurately localized, but predicted as \textit{bouquet}.

\noindent {\bf Separating the prior shift} 
\tk{
To further delineate prior shift from context and design shifts, we curate a balanced subset out of \Ours{} such that each category has about 200-300 images, and drop categories which have fewer images (about $3/4^{th}$ of the categories remain). From \cref{tab:balanced-acc}, the drop in accuracy after addressing the prior shift is 12.9\% on \GeoP{} and 15.4\% on \GeoI{}, compared to 20.08\% and 19.37\% on the original datasets, showing that non-trivial accuracy drops caused by context and design shifts still exist even after accounting for label imbalance between the domains. 
}

\subsection{Benchmarking Domain Adaptation}
\label{sec:exp_uda}

\noindent We study the effectiveness of prior unsupervised adaptation algorithms in bridging novel notions of domain gaps like context shift and design shift on \Ours{}. We review various standard as well as current state-of-the-art domain adaptation methods to examine their geographical robustness. 

\vspace{4pt}
\noindent {\bf Architecture and training details} We follow the standard protocol established in prior works~\cite{CDAN, saito2018maximum, kalluri2022memsac} and use an ImageNet pre-trained Resnet-50~\cite{he2016deep} as the feature extractor backbone and a randomly intialized classifier layer. 
We use a batch size of 32 and SGD with a learning rate of 0.01 for the classifier head and 0.001 for the already pretrained backbone. We report the top-1 and top-5 accuracy numbers using the test splits from each benchmarks. 
We perform comparisons between traditional adversarial methods (DANN~\cite{DANN}, CDAN~\cite{CDAN}), class-aware adaptation methods (MCC~\cite{jin2020minimum}, MDD~\cite{zhang2019bridging}), non-adversarial methods (SAFN~\cite{xu2019larger}, MCD~\cite{saito2018maximum}) as well as recent state-of-the-art (ToAlign~\cite{wei2021toalign}, MemSAC~\cite{kalluri2022memsac}). We train prior works using their publicly available code and adopt all hyper-parameters as recommended in the respective papers. 

\vspace{2pt}
\noindent {\bf Existing UDA methods do not suffice on \Ours{}} We show the Top-1 and Top-5 accuracies of all the transfer settings from \Ours{} in \cref{tab:uda-benchmarking}. A key observation is that most of the domain adaptation approaches are no better, or sometimes even worse, than the baseline model trained only using source domain data, indicating their limitations for geographic domain adaptation. For example, on \GeoP{}, training using data from USA achieves a top-1 accuracy of 36.27\% on test data from Asia test images, while the best adaptation method (MemSAC) obtains lesser accuracy of 34.7\%, indicating negative transfer. Likewise, on \GeoI{}, a USA-trained source model achieves 36.98\% on test images from Asia which is comparable to the best adaptation accuracy of {36.71}\%. To further illustrate this, we define relative accuracy gain as the improvement in accuracy obtained by a method over a source-only model as a percentage of gap between a source-only model and the target-supervised upper bound (which is 100\% if the method achieves the target supervised upper bound).
From \cref{fig:teaser_adaptation}, it is notable that the same adaptation methods that yield significantly high relative accuracy gains on DomainNet~\cite{peng2019moment} yield negative relative accuracy gains on \Ours{}, highlighting the unique the nature of distribution shifts in real-world settings like geographic adaptation that challenge existing methods. 
These observations also suggest that future research should focus on context-aware and object-centric representations in addition to domain invariant features to improve cross-domain transfer amidst context and design shifts.

% \vspace{2pt}
\noindent {\bf Universal domain adaptation on Geo-UniDA} We run SOTA universal domain adaptation methods (You et.al.~\cite{you2019universal}, DANCE~\cite{saito2020dance} and OvaNET~\cite{saito2021ovanet}) on the Geo-UniDA benchmark of \Ours{}. 
Following prior works~\cite{saito2021ovanet}, we adopt the H-score metric which is a harmonic mean of closed-set and open-set accuracies giving equal importance to closed set transfer as well as open set accuracy. In \cref{tab:uniDA-benchmarking}, we show that DANCE~\cite{saito2020dance} outperforms both You et.al.~\cite{you2019universal} and OVANet~\cite{saito2021ovanet} on the Geo-UniDA benchmark. We also show that a significant gap still exists between target supervised accuracy when trained using supervision (70.7\%) and best adaptation accuracy (38.5\%) on our benchmark, highlighting the limitations of existing methods to efficiently address universal adaptation in a geographic context.

\subsection{Large-scale pre-training and architectures} 
\label{subsec:pretraining_Exp}

\noindent It is common to use large scale self-supervised~\cite{he2020momentum, caron2020unsupervised, chen2020simple, chen2021empirical, caron2021emerging, he2022masked} and weakly-supervised~\cite{jia2021scaling, singh2022revisiting, mahajan2018exploring} pre-trained models as starting points in various downstream applications. While recent works explored role of pre-training on domain robustness~\cite{kim2022broad}, we are interested in the extent to which large scale pre-training effectively preserved robustness when fine-tuned on geographically under-represented datasets. We investigate the performance of a variety of methods on \Ours{} in terms of backbone architectures, pre-training strategies and supervision.

%%%%%%%%%%%%%%%%%%%%%%%%%%%%%%%%
%%%%%%%%%%%%%%%%%%%%%%%%%%%%%%%%
\begin{figure}[t]
\begin{center}
   \centering
        \includegraphics[width=0.47\textwidth]{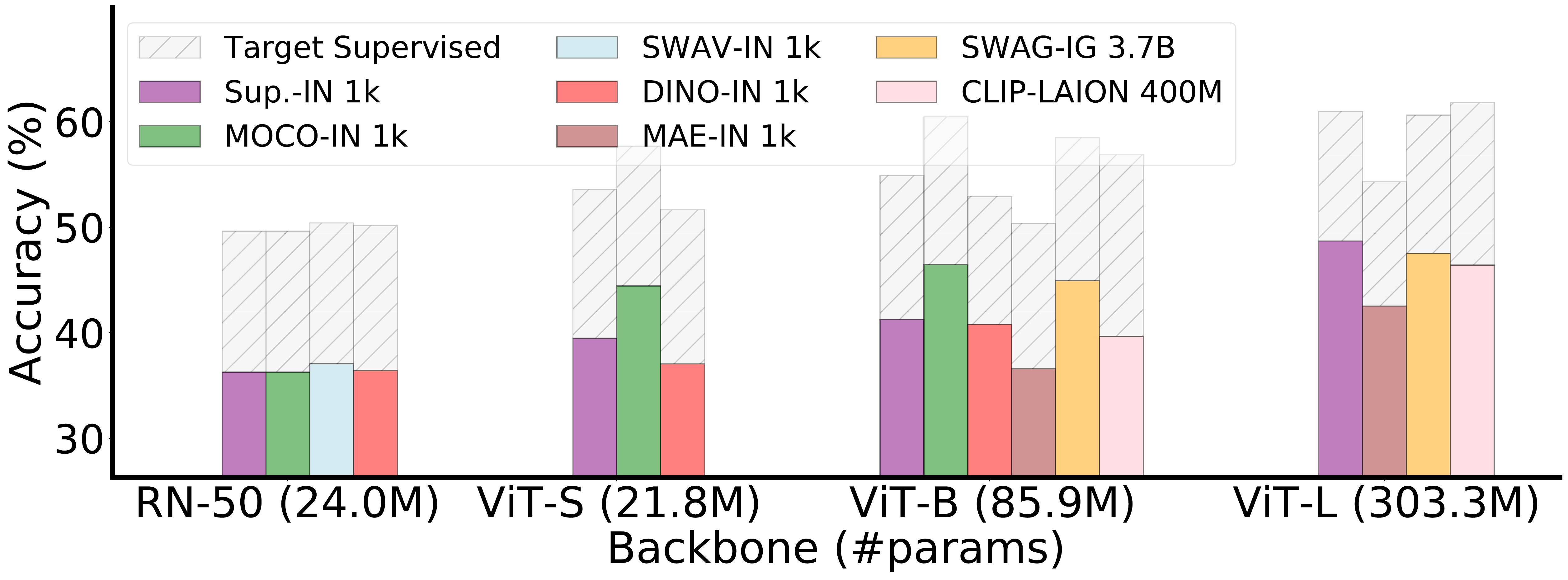}
        \vspace{-9pt}
        \caption{\tk{We show that most architectures and pre-training strategies exhibit significant cross-domain drops when fine-tuned on geographically biased datasets. Shown for USA$\rightarrow$Asia on \GeoP{}, refer \cref{fig:teaser_archs} for the plot on \GeoI{} and \supp{} for other transfer settings.}}
        \label{fig:vit-places}
        \vspace{-18pt}
\end{center}
\end{figure}
%%%%%%%%%%%%%%%%%%%%%%%%%%%%%%%%
%%%%%%%%%%%%%%%%%%%%%%%%%%%%%%%%

\vspace{4pt}
\noindent {\bf Experimental setup} Our backbone architectures include \rfy{}~\cite{he2016deep} as well as the small (\vits{}), base (\vitb{}) and large (\vitl{}) vision transformers~\cite{dosovitskiy2020image}. In terms of supervision, in addition to the standard supervised pre-training on ImageNet-1k, we also consider self-supervised methods MoCo-V3~\cite{chen2021empirical}, SwAV~\cite{caron2020unsupervised}, DINO~\cite{caron2021emerging}, MAE~\cite{he2022masked} trained on ImageNet-1k, the weakly supervised SWAG~\cite{singh2022revisiting} trained on 3.6B uncurated instagram images and CLIP~\cite{radford2021learning} trained on 400M image-language pairs~\cite{schuhmann2021laion}. We denote \{Backbone-Supervision-Data\} for different model choices (for example, \noop{Resnet50-sup-IN1k} indicates a \rfy{} pre-trained on supervised data from ImageNet-1k).

For evaluating geographic robustness of these models, we first take the pre-trained model and fine-tune it on training data from a ``source'' geography, then evaluate the performance on test data from the ``target'' geography. We show the results using USA as the source and Asia as the target from the \GeoP{} benchmark in \cref{fig:vit-places}, and \GeoI{} benchmark in \cref{fig:teaser_archs}. 
For reference, we also report accuracy after fine-tuning on labeled data from the target geography for each \{Backbone-Supervision-Data\} pair (denoted as {target-supervised}), which serves as an upper bound for the transfer performance. 

\vspace{4pt}
\noindent {\bf Large-scale pretraining is not geographically robust} %
From \cref{fig:vit-places}, we make a few observations. 
Firstly, comparison between \rfy{} and \vits{} which have roughly the same number of parameters suggests the superiority of the vision transformer architectures over CNNs. For example, \noop{\vits{}-sup-IN1k} is better than \noop{\rfy{}-sup-IN1k}, and \noop{\vits{}-moco-IN1k} is better than \noop{\rfy{}-moco-IN1k}, indicating that global reasoning using self-attention layers in vision transformers benefits context-dependent tasks like \GeoP{}. 
Next, comparing different pre-training strategies, we observe that MoCo gives best accuracy on \vits{} and \vitb{}, while supervised pre-training outperforms other approaches on large models like \vitl{}.
However, the gap between target supervised accuracy and the best adaptation accuracy achieved using either \rfy{} or any of the vision transformers is still high, highlighting the need for better transfer strategies. 
In terms of data, weakly-supervised pre-training using billion-scale dataset IG3.6B (\noop{\vitb{}-swag-3B}) shows significant improvements over self-supervised training methods like MAE (\noop{\vitb{}-mae-IN1k}) and DINO (\noop{\vitb{}-dino-IN1k}). But despite training on massive-scale data, \noop{\vitl{}-swag-3B} and {\vitl{}-clip-400M} are still inferior to the target supervised accuracies, revealing the limitations of current pre-training strategies towards robust cross-geography transfer after fine-tuning. While the success of large-scale pre-training strategies are well-documented on popular datasets like ImageNet, our results indicate that similar benefits might not be observed when application domains significantly differ from pre-training or fine-tuning datasets~\cite{DBLP:conf/cvpr/ColeYWAB22}. 

%%%%%%%%%%%%%%%%%%%%%%%%%%%%%%%%%%%%%%%%%%%%%%%%%%%%%%%%%%%%%%
%%%%%%%%%%%%%%%%%%%%%%%%%%%%%%%%%%%%%%%%%%%%%%%%%%%%%%%%%%%%%%

\section{Conclusion}
\label{sec:conclusion}

We introduce a new dataset called \Ours{} for the problem of geographic adaptation with benchmarks covering the tasks of scene and object classification. In contrast to existing datasets for domain adaptation~\cite{peng2017visda, peng2019moment, office31, venkateswara2017deep}, our dataset with images collected from different locations contains domain shifts captured by natural variations due to geographies, cultures and weather conditions from across the world, which is a novel and understudied direction in domain adaptation. Through \Ours{}, we analyze the sources of domain shift caused by changes in geographies such as context and design shift. We conduct extensive benchmarking on \Ours{} and highlight the limitations of current domain adaptation methods as well as large-scale pretraining methods towards geographical robustness. 
Finally, in spite of geographical diversity in \Ours{}, we note a possible limitation of indirect bias towards USA as the user-base on photo-sharing sites like Flickr is dominated by the US. Creating datasets that are a more natural reflection of cultures and trends from diverse geographies and devising learning algorithms robust to those variations is an exciting proposition for the future.

\noindent {\bf Acknowledgements} We thank NSF CAREER 1751365, Google AI Award for Inclusion Research and National Research Platform for hardware access.

%%%%%%%%%%%%%%%%%%%%%%%%%%%%%%%%%%%%%%%%%%%%%%%%%%%%%%%%%%%%%%
%%%%%%%%%%%%%%%%%%%%%%%%%%%%%%%%%%%%%%%%%%%%%%%%%%%%%%%%%%%%%%

%\newpage
%%%%%%%%% REFERENCES
{\small
\bibliographystyle{ieee_fullname}
\bibliography{main}
}

\begin{appendix}
\appendix
\section{Performance on additional geographies}
In Table 2 in the main paper, we illustrated cross-domain drops across geographies for the case of USA$\leftrightarrow$Asia. We show that this phenomenon is not specific to these geographies, and similar cross-domain drop in accuracy can be observed in case of Africa as a new geographical domain. For this purpose, we follow a similar pipeline discussed in Section 3.1 of the main paper and collect images from Africa belonging to the 205 classes from Places-205, creating the test-set for Africa domain for \GeoP{} with 8358 images. 
% However, we couldn't conduct a similar exercise for \GeoI{} as WebVision~\cite{li2017webvision}, the original source of \GeoI{}, does not have enough images from Africa from the 700 concepts we selected. Nevertheless, 
For the case of \GeoP{}, we show in \cref{tab:source-transfer-africa} that a model trained on USA obtains only $32.2\%$ on test images from Africa with a significant drop of $24\%$, and a model trained on images from Asia only gets $26.77\%$ top-1 accuracy on Africa test images with a drop of $23\%$ compared to within-domain test accuracy. These results indicate that cross-domain transfer exhibits similar challenges across any geographically separated domains.

\begin{table}[!t]
  \centering
  \resizebox{0.46\textwidth}{!}{
  % \begin{tabular}{@{} >{\raggedright}p{5.2cm} >{\raggedright}p{2.2cm} >{\raggedright}p{2.2cm} >{\raggedright}p{2cm} p{2cm}}
  \begin{tabular}{lccccccccc}
    & \multicolumn{3}{c}{\textbf{\underline{\GeoP}}}  \\
    Train $\downarrow$ / Test $\rightarrow$ & USA & Asia & Africa \\
    \cline{2-4}
    USA     & 56.35/85.15 & 36.27/63.27 & 32.20/51.97 \\
    Asia    & 21.03/44.81 & 49.63/78.45 & 26.77/47.90 \\
    % Africa  & 9.18/21.5   & 13.95/31.13 & 46.16/68.01 \\
    % \textcolor{gray}{USA+Asia} & \textcolor{gray}{yy.yy/yy.yy} & xx.xx/xx.xx &  \\
    % \midrule 
    % & \multicolumn{3}{c}{\textbf{\underline{\GeoI}}} \\
    % Train $\downarrow$ / Test $\rightarrow$ & USA & Asia & Drop (\%) \\
    % \cline{2-4}
    % USA & 46.63/67.85 & 29.69/51.43 & \textcolor{brown}{-16.94/-16.42} \\
    % Asia & 31.55/52.28 & 52.93/72.96 & \textcolor{brown}{-21.38/-20.68} \\
    % \textcolor{gray}{USA+Asia} & \textcolor{gray}{yy.yy/yy.yy} & \textcolor{gray}{xx.xx/xx.xx} &  \\
    \bottomrule
  \end{tabular}
  }
  \vspace{-1em}
  \captionsetup{width=0.45\textwidth}
  \caption{\label{tab:source-transfer-africa} {\bf Cross-Geography Drops on \GeoP{}} Top-1/Top-5 accuracies of Resnet-50 models across geographically different train and test domains, including a new test-set from Africa domain.}
  \vspace{-1em}
\end{table}
%%%%%%%%%%%%%%%%%%%%%%%%%%%%%%%%%%%%%%%%%%%%%%%%%%%
%%%%%%%%%%%%%%% tSNE of context and designs %%%%%%%
%%%%%%%%%%%%%%%%%%%%%%%%%%%%%%%%%%%%%%%%%%%%%%%%%%%%
\begin{figure*}[t]
\begin{center}
    \begin{minipage}[b]{0.23\textwidth}
        \centering
        \includegraphics[width=\textwidth]{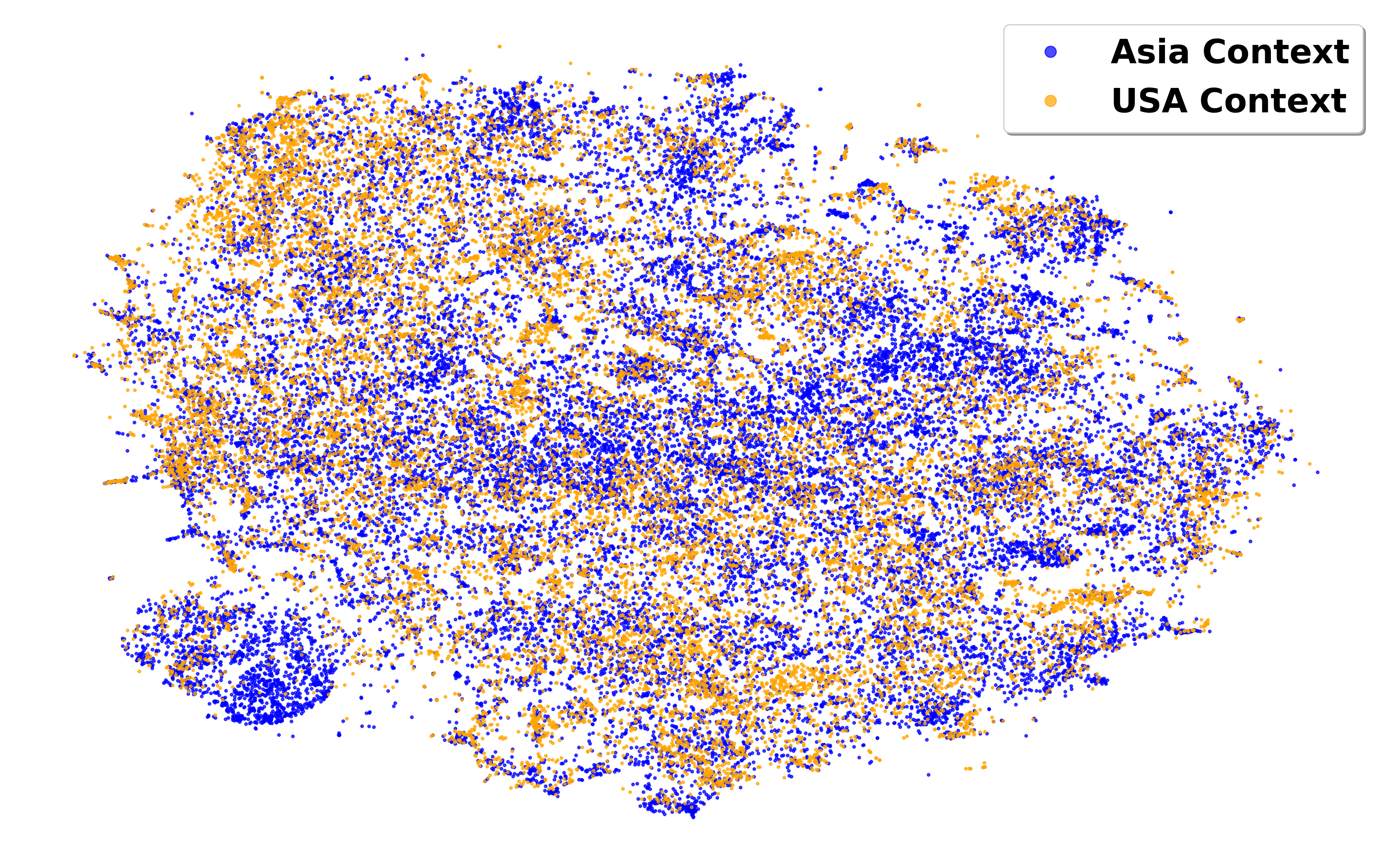}
        \vspace{-16pt}
        \subcaption{{\bf Context Shift in \GeoP{}} }
        \label{fig:tsne_context_places}
    \end{minipage}
    \hfill
    \begin{minipage}[b]{0.23\textwidth}
        \centering
        \includegraphics[width=\textwidth]{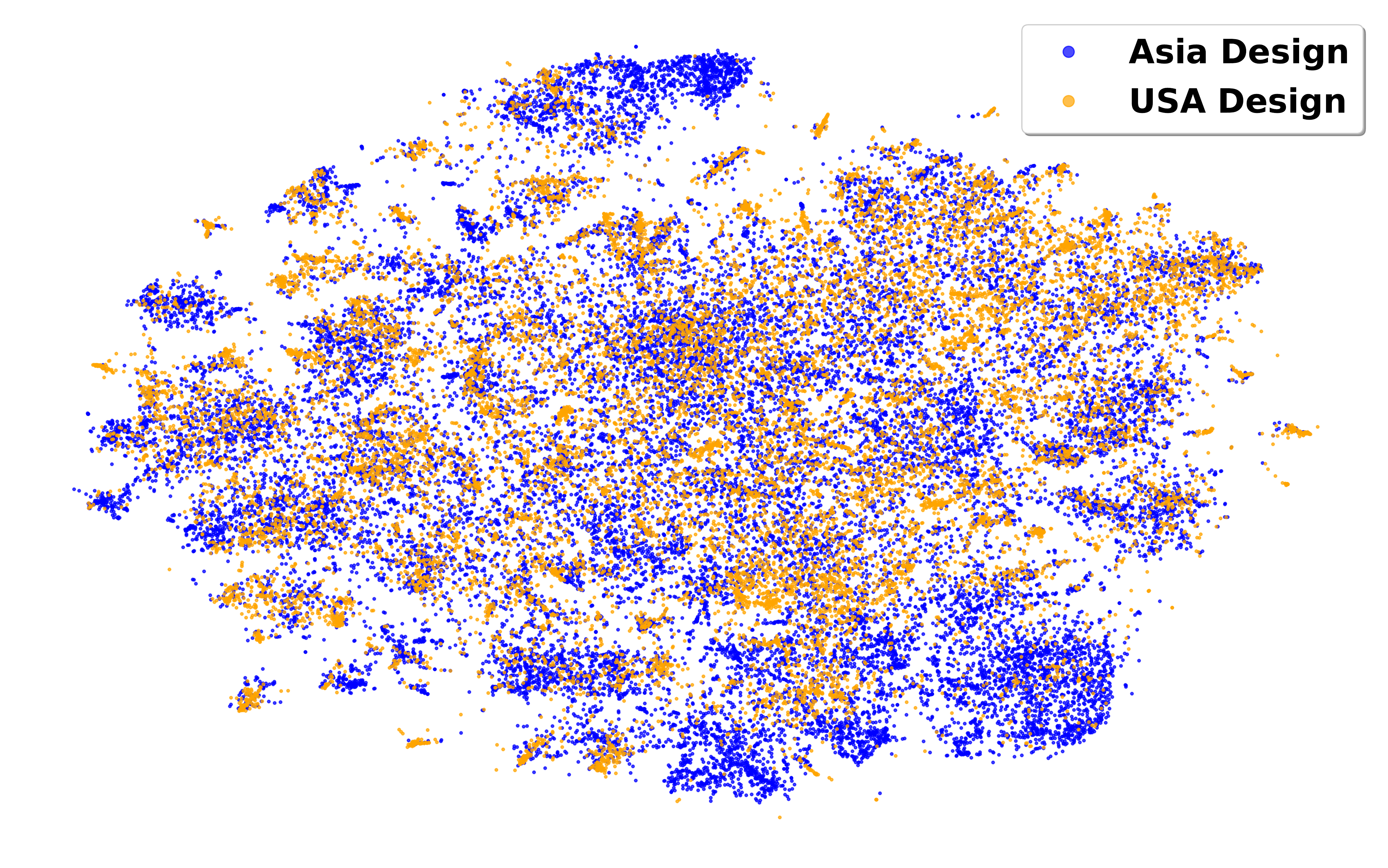}
        \vspace{-16pt}
        \subcaption{{\bf Design Shift in \GeoP{}} }
        \label{fig:tsne_design_places}
    \end{minipage}
    \hfill
    \begin{minipage}[b]{0.23\textwidth}
        \centering
        \includegraphics[width=\textwidth]{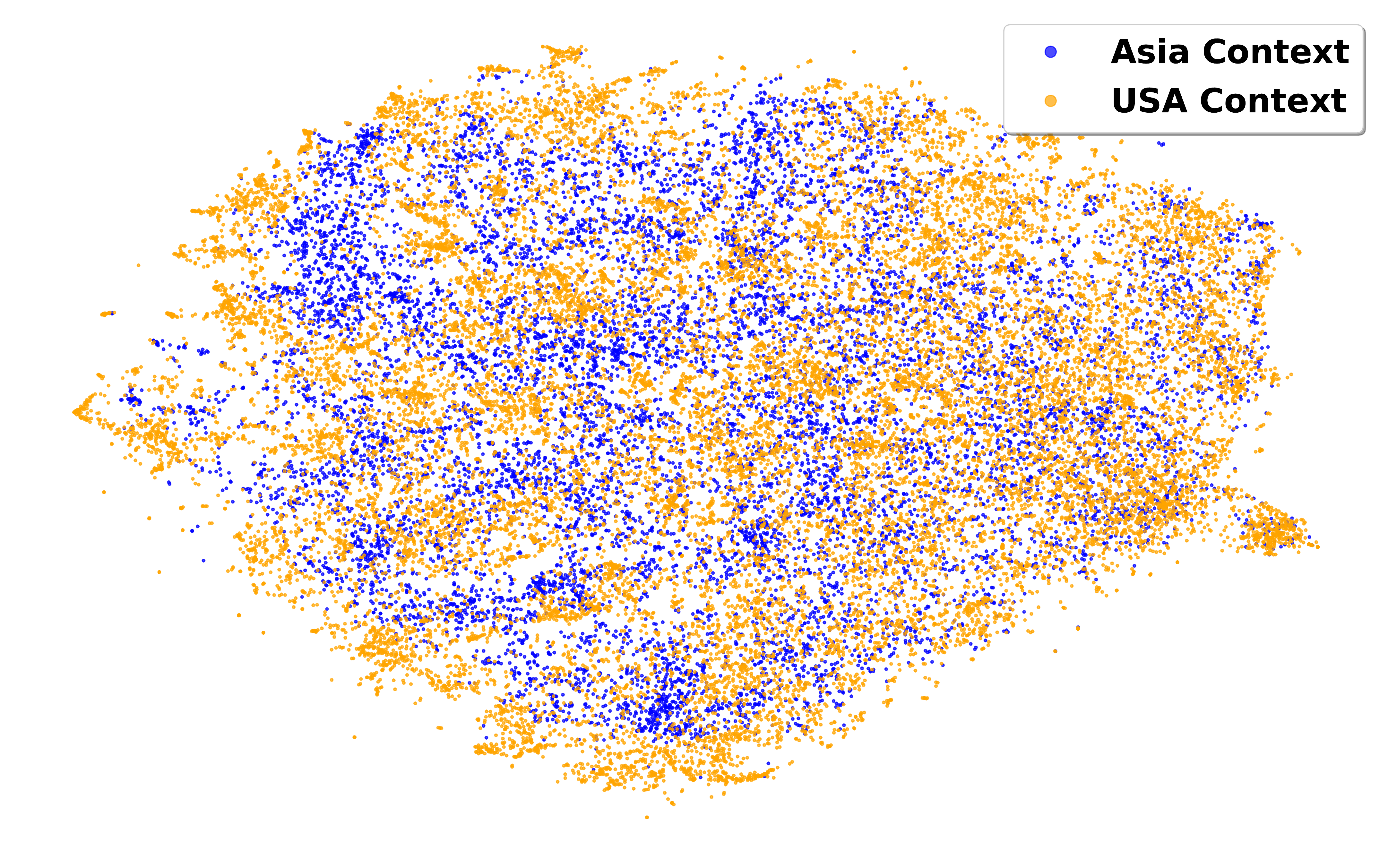}
        \vspace{-16pt}
        \subcaption{{\bf Context Shift in \GeoI{}} }
        \label{fig:tsne_context_imnet}
    \end{minipage}
    \hfill
    \begin{minipage}[b]{0.23\textwidth}
        \centering
        \includegraphics[width=\textwidth]{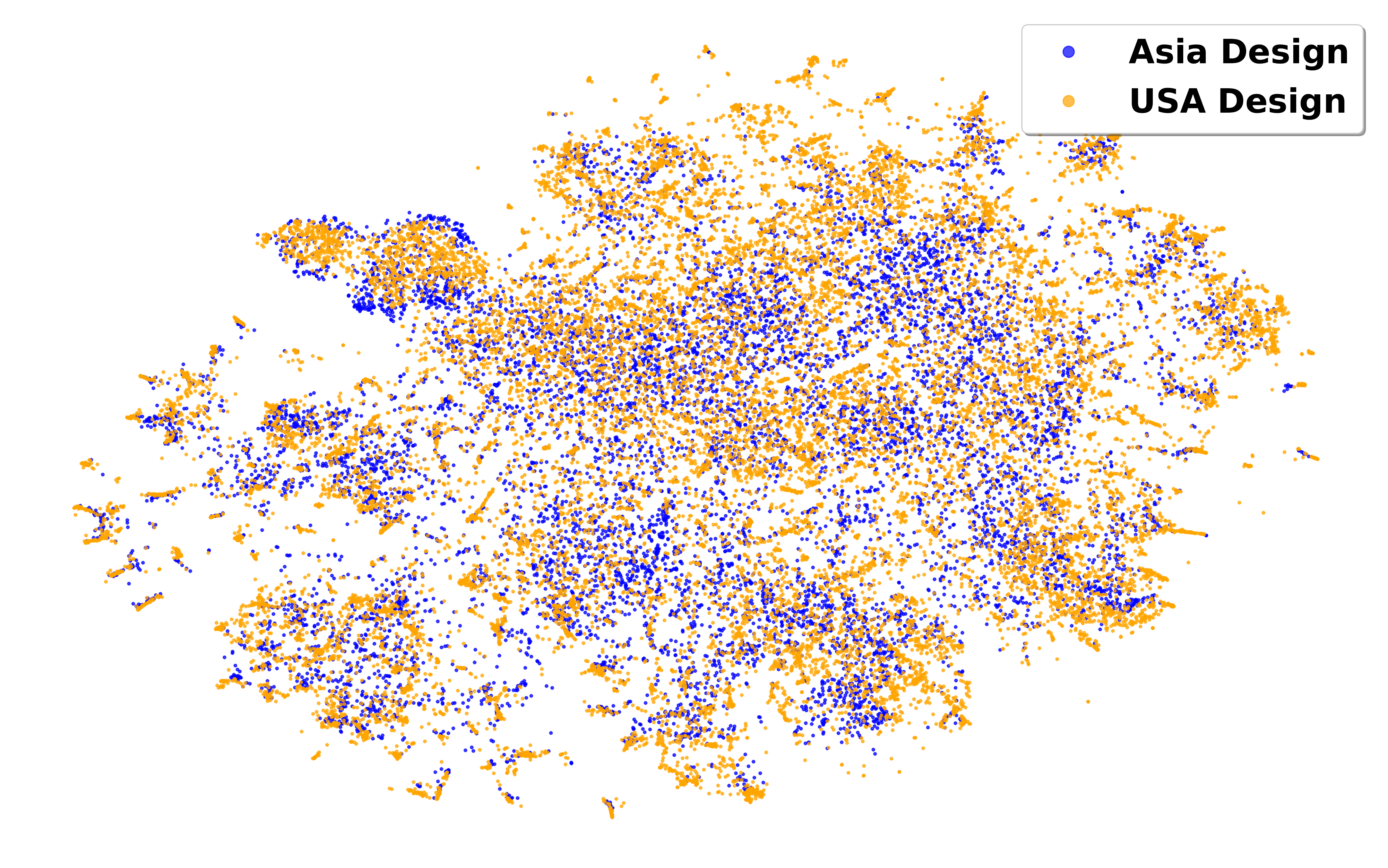}
        \vspace{-16pt}
        \subcaption{{\bf Design Shift in \GeoI{}} }
        \label{fig:tsne_design_imnet}
    \end{minipage}

    % \vspace{-0.1cm}
    \captionsetup{width=\textwidth, font=footnotesize}
    \caption{{\bf tSNE Visualizations of context and design shifts in \Ours{}}. As shown, there is a notable separation between the context and design features between USA (in orange) and Asia (in blue) in both \GeoP{} and \GeoI{}.}
    \label{fig:tSNE}
    % \vspace{-3em}
\end{center}
\end{figure*}

\section{Visualization of Context and Design Shifts}
We provide deeper insight into the cross-domain shifts in contexts and designs induced by the geographies by visualizing their tSNE feature representations~\cite{van2008visualizing}. To this end, we first recall that
% definitions of the \textit{context} and \textit{design} from Section 3.4 in the main paper. Specifically, 
we defined context of an image $x$ as $b_x$ representing the background regions in an image, and design $f_x$ as the foreground objects (Section 3.4 in the main paper). However, we do not have box or mask annotation corresponding to the images in \Ours{}, so it is not possible to directly infer the context and foreground in each image. Instead, we rely on a state-of-the-art object detector Mask-RCNN trained on COCO dataset~\cite{lin2014microsoft} for this purpose. Specifically, we train a class-agnostic Mask-RCNN on the COCO dataset by mapping all the class labels to a single foreground class.
We then identify all the masks detected by the network on our images, so that these masks then correspond to the foreground objects, while the other parts of the image corresponds to the background. To compute the feature representation of the foreground objects, we element-wise multiply the binary foreground mask with the deep feature map from the backbone Resnet-50, followed by a global pool. In other words, we use the binary foreground mask to select the area from the feature map corresponding to the foreground, and take an average of the locations to obtain a 2048-dimensional foreground feature vector per image. We similarly obtain a 2048-dimensional background vector by using the negation of the binary foreground mask as the background mask. Therefore, we end up with two feature representations per image pertaining to the foreground (design) and background (context) respectively. We repeat this for both domains USA and Asia from both the \GeoP{} and \GeoI{} splits of our dataset. We then project this 2048 dimensional vector into a 2-dimensional vector using tSNE reduction and visualize the embeddings in \cref{fig:tSNE}. 

\noindent {\bf Context Shift} The pronounced distinction in the contexts between the two domains from \GeoP{} is highlighted in \cref{fig:tsne_context_places}, where we show minimum overlap between the features corresponding to the background regions in USA and Asia. Similar observations also hold for the case of \GeoI{} in \cref{fig:tsne_context_imnet}. Since the background or the context plays a major role in identifying places or objects, this shift invariably results in drop in accuracy under cross-geography transfer.

%%%%%%%%%%%%%%%%%%%%%%%%%%%%%%%%%%%%%%%%%%%%%%%%%%%
%%%%%%%%%%%%%%% Geographic Distribution %%%%%%%%%%%
%%%%%%%%%%%%%%%%%%%%%%%%%%%%%%%%%%%%%%%%%%%%%%%%%%%
\begin{figure*}[t]
\begin{center}
    \begin{minipage}[b]{0.48\textwidth}
        \centering
        \includegraphics[width=\textwidth]{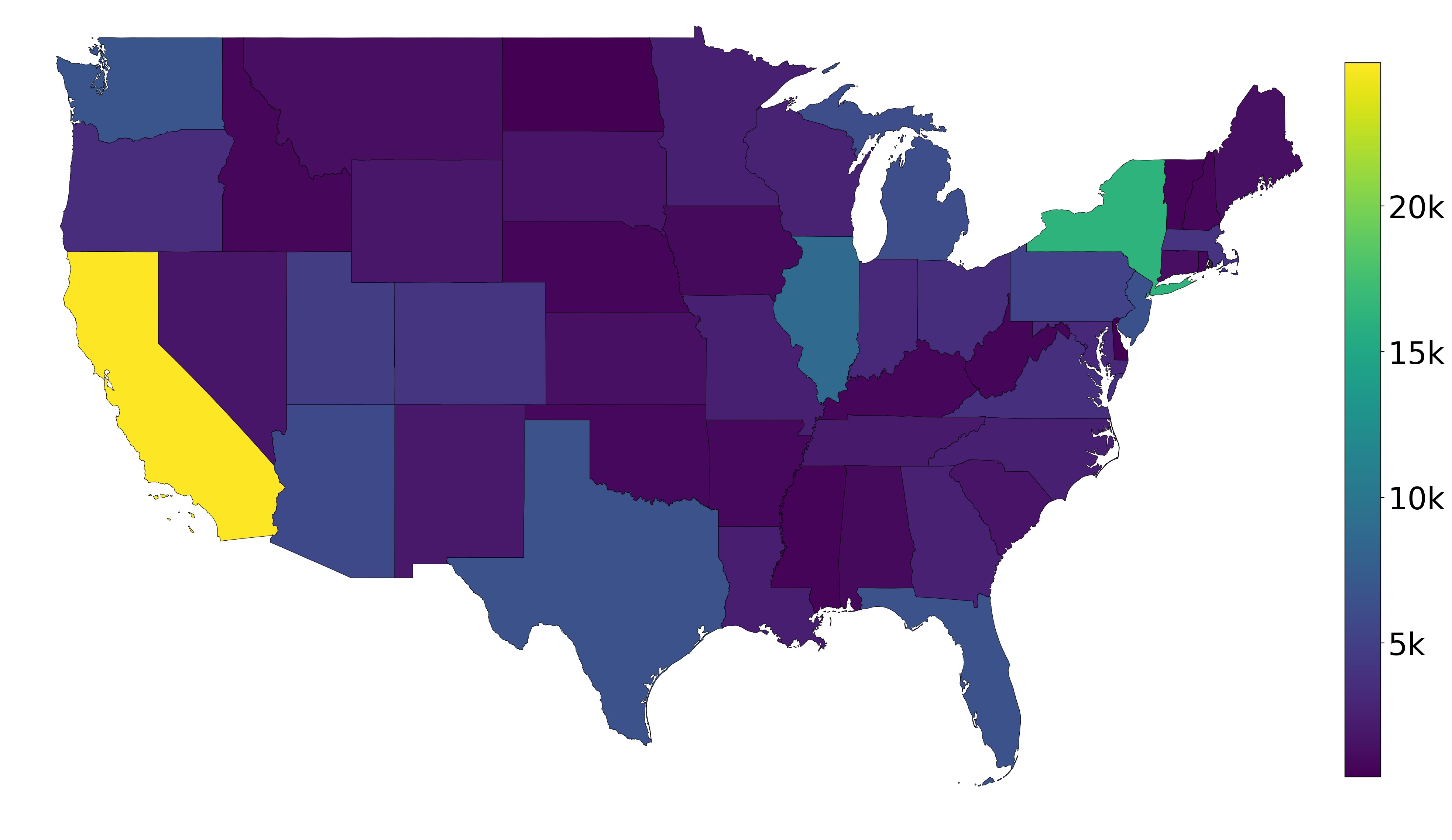}
        % \vspace{-8pt}
        \subcaption{{\bf \GeoP{}: USA Images} }
        \label{fig:map_places_usa}
    \end{minipage}
    \hfill
        \begin{minipage}[b]{0.48\textwidth}
        \centering
        \includegraphics[width=\textwidth]{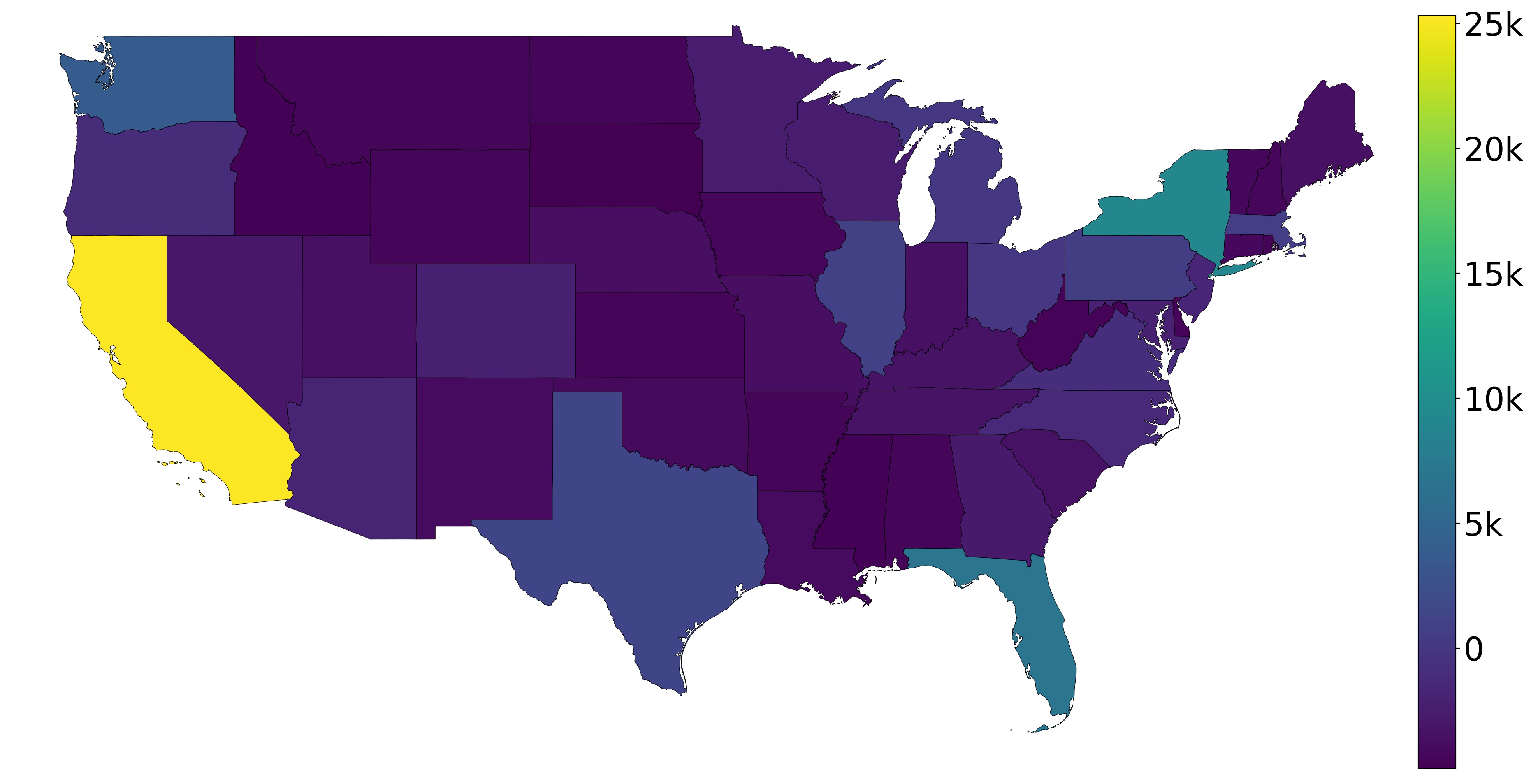}
        % \vspace{-8pt}
        \subcaption{{\bf \GeoI{}: USA Images} }
        \label{fig:map_imnet_usa}
    \end{minipage}
    
    \begin{minipage}[b]{0.48\textwidth}
        \centering
        \includegraphics[width=\textwidth]{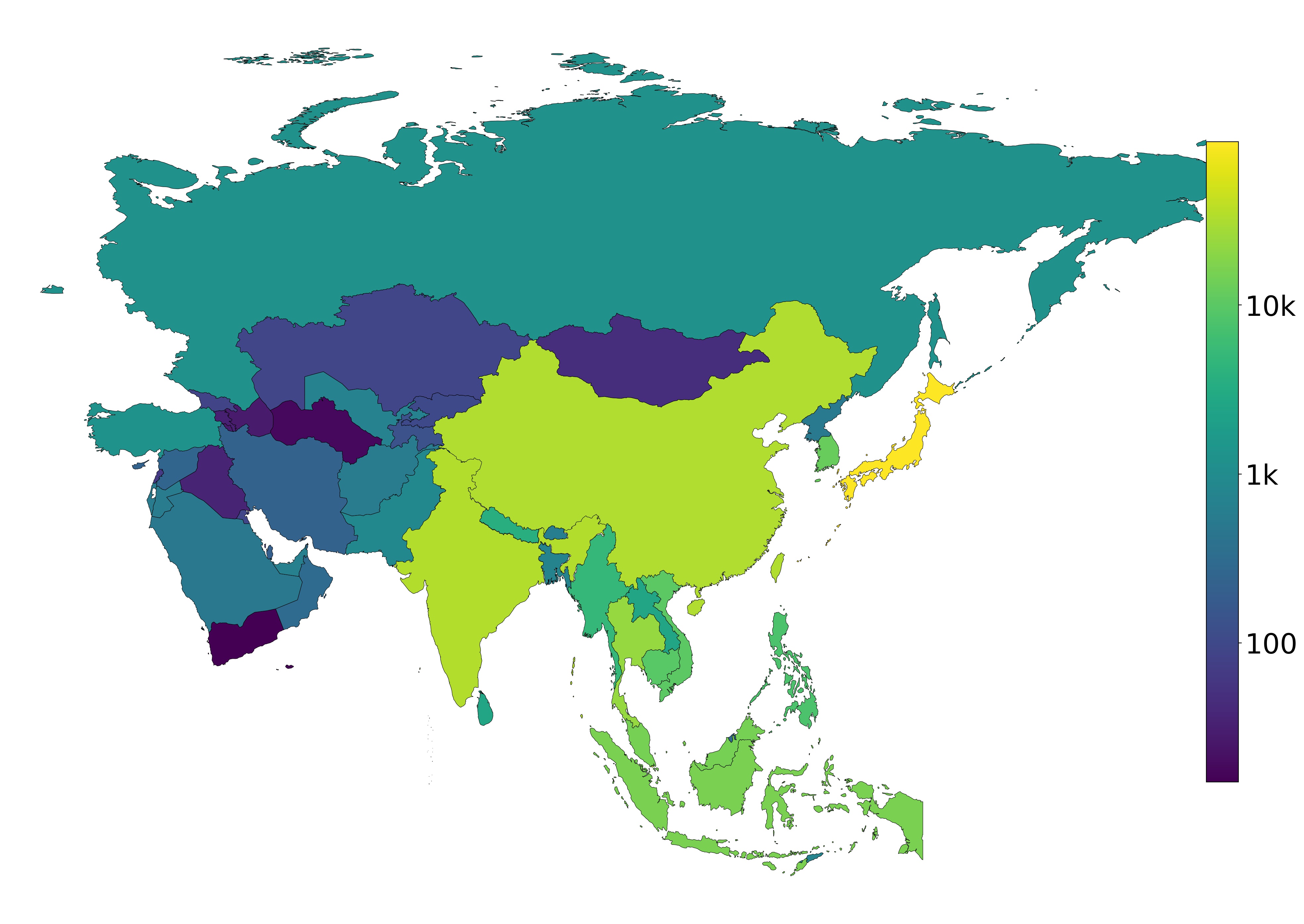}
        % \vspace{-8pt}
        \subcaption{{\bf \GeoP{}: Asia Images} }
        \label{fig:map_places_asia}
    \end{minipage}
    \hfill
    \begin{minipage}[b]{0.48\textwidth}
        \centering
        \includegraphics[width=\textwidth]{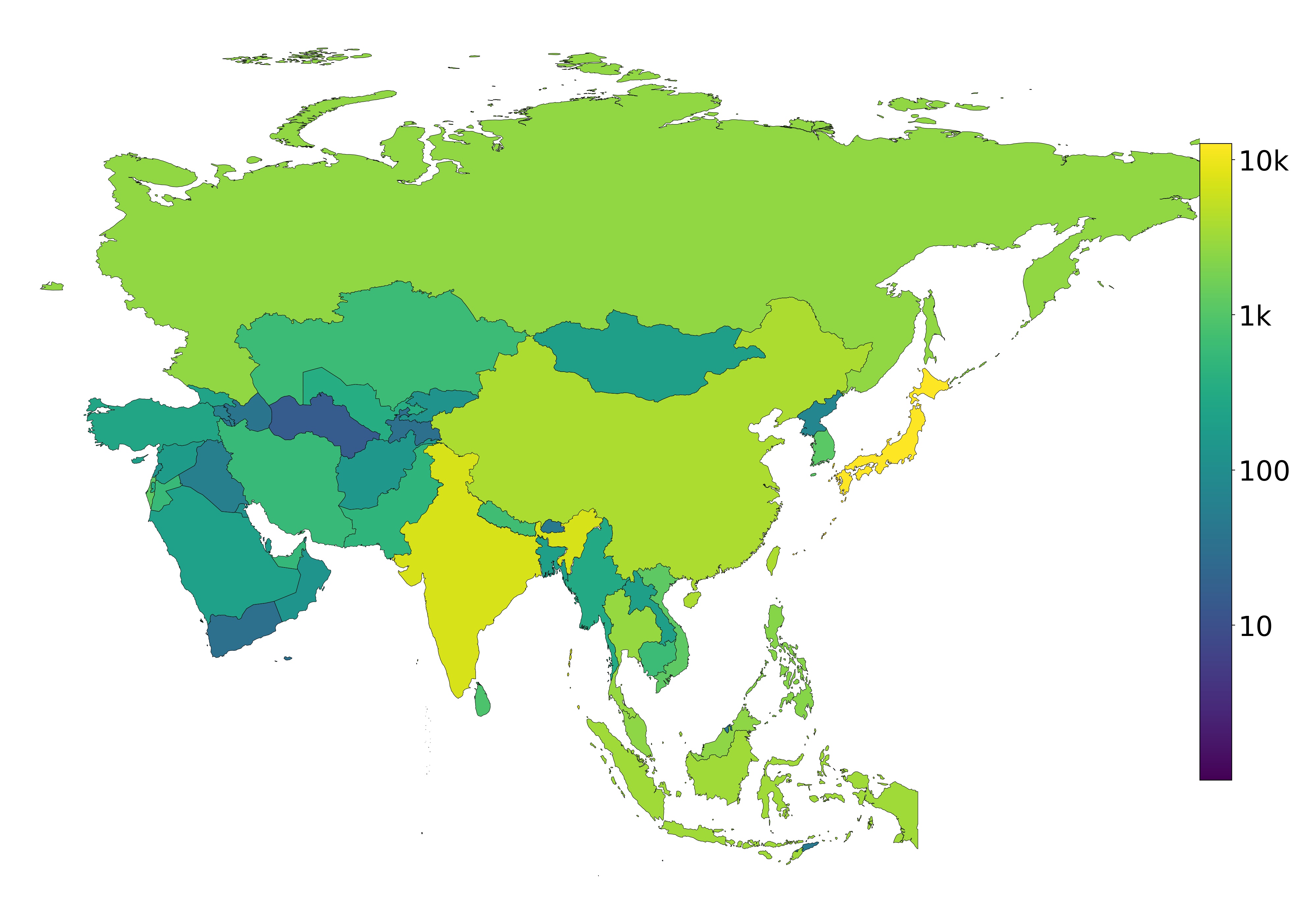}
        % \vspace{-8pt}
        \subcaption{{\bf \GeoI{}: Asia Images}}
        \label{fig:map_imnet_asia}
    \end{minipage}
    
    \vspace{-0.1cm}
    \captionsetup{width=\textwidth, font=footnotesize}
    \caption{{\bf Geographical Distribution of images from USA and Asia domains}. We show the images per geographical sub-region in both domains on \Ours{}. As shown, in Asia, a majority of images are from Japan, India, Korea, China and Taiwan while in USA, a majority of images are from populous regions like California and New York. Note that the color-bar scale is linear for USA and log-scale for Asia.}
    \label{fig:worldmaps_dist}
\end{center}
\end{figure*}
%%%%%%%%%%%%%%%%%%%%%%%%%%%%%%%%%%%%%%%%%%%%%%%%%%%%%%%%%%
%%%%%%%%%%%%%%%%%%%%%%%%%%%%%%%%%%%%%%%%%%%%%%%%%%%%%%%%%%

\noindent {\bf Design Shift} The tSNE features of the foreground regions is shown in \cref{fig:tsne_design_places} for the case of \GeoP{} and in \cref{fig:tsne_design_imnet} for \GeoI{}. Minimum overlap is observed between the features corresponding to the foreground, or design of the objects, in each case indicating the presence of notable design shift between the domains. % which also shown in the poor cross-geography transfer.

We also note that datasets like COCO are predominantly US-biased, so the use of COCO in analyzing distribution shifts on Asia images is not completely fair. To this end, manually annotating images with finer-grained foreground and context labels in both geographies would yield more accurate analysis, which is left as a future work.

%%%%%%%%%%%%%%%%%%%%%%%%%%%%%%%%%%%%%%%%%%%%%%%%%%%
%%%%%%%%%%%%%%% Unsupervised Adaptation %%%%%%%%%%%
%%%%%%%%%%%%%%%%%%%%%%%%%%%%%%%%%%%%%%%%%%%%%%%%%%%
\begin{figure*}[t]
\begin{center}
    \begin{minipage}[b]{0.9\textwidth}
        \centering
        \includegraphics[width=\textwidth]{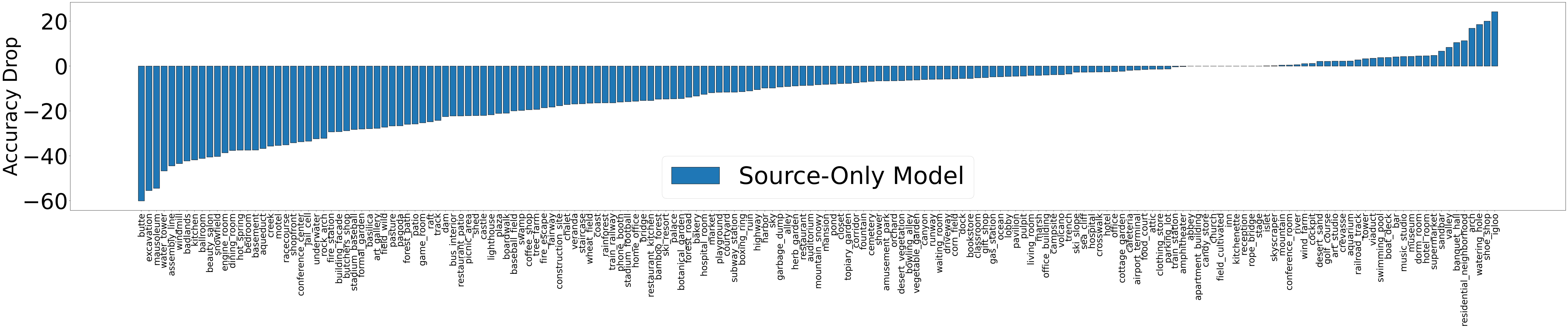}
        \vspace{-16pt}
        \subcaption{{\bf Source-Only Training} }
        \label{fig:da_plain}
    \end{minipage}

     \vspace{.2cm}
     \begin{minipage}[b]{0.9\textwidth}
        \centering
        \includegraphics[width=\textwidth]{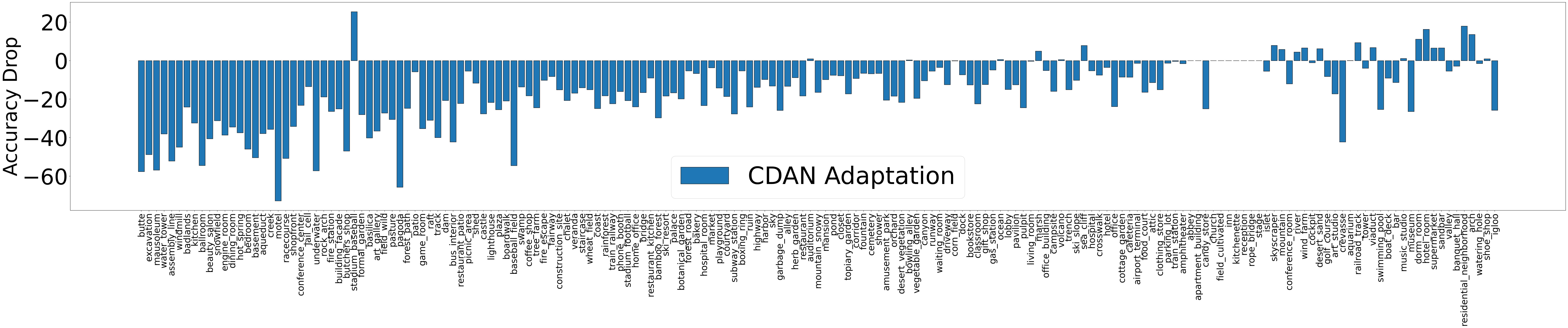}
        \vspace{-16pt}
        \subcaption{{\bf CDAN Adaptation} }
        \label{fig:da_cdan}
    \end{minipage}

    \vspace{.2cm}
    \begin{minipage}[b]{0.9\textwidth}
        \centering
        \includegraphics[width=\textwidth]{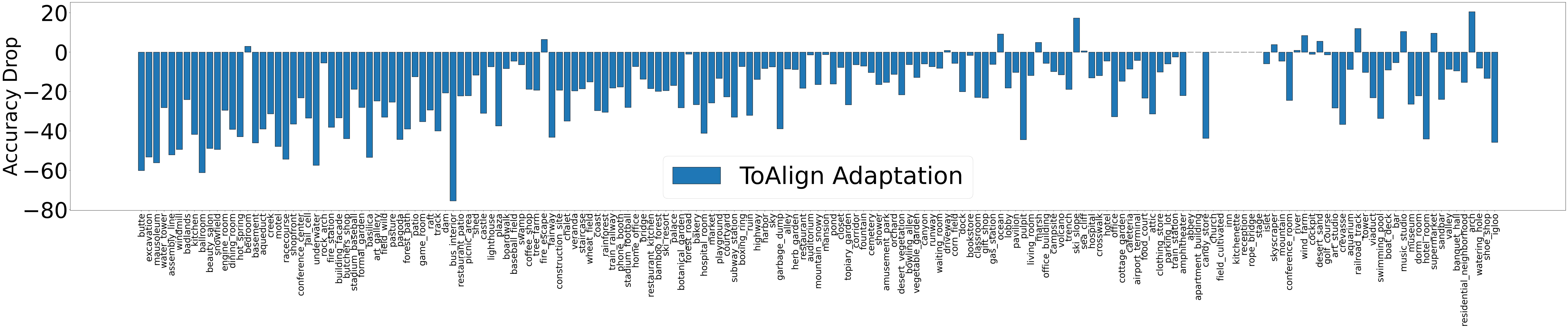}
        \vspace{-16pt}
        \subcaption{{\bf ToAlign Adaptation} }
        \label{fig:da_toalign}
    \end{minipage}
    
    \vspace{.2cm}
    \captionsetup{width=\textwidth, font=footnotesize}
    \caption{{\bf Per-class accuracy drops} on USA$\rightarrow$Asia transfer for a plain source-only model as well as post-adaptation using CDAN~\cite{CDAN} and ToAlign~\cite{wei2021toalign} adaptation methods. Note that the trend of per-class accuracy drops is the same before and after the adaptation indicating the limited benefit offered by existing state-of-the-art adaptation methods in bridging geographical shifts.} 
    \label{fig:perclass_accuracies}
\end{center}
\end{figure*}
%%%%%%%%%%%%%%%%%%%%%%%%%%%%%%%%%%%%%%%%%%%%%%%%%%%%%%%%%%
%%%%%%%%%%%%%%%%%%%%%%%%%%%%%%%%%%%%%%%%%%%%%%%%%%%%%%%%%%

%%%%%%%%%%%%%%%%%%%%%%%%%%%%%%%%%%%%%%%%%%%%%%%%%%%
%%%%%%%%%%%%%%% Large-scale training    %%%%%%%%%%%
%%%%%%%%%%%%%%%%%%%%%%%%%%%%%%%%%%%%%%%%%%%%%%%%%%%
\begin{figure*}[t]
\begin{center}
    \begin{minipage}[b]{0.45\textwidth}
        \centering
        \includegraphics[width=\textwidth]{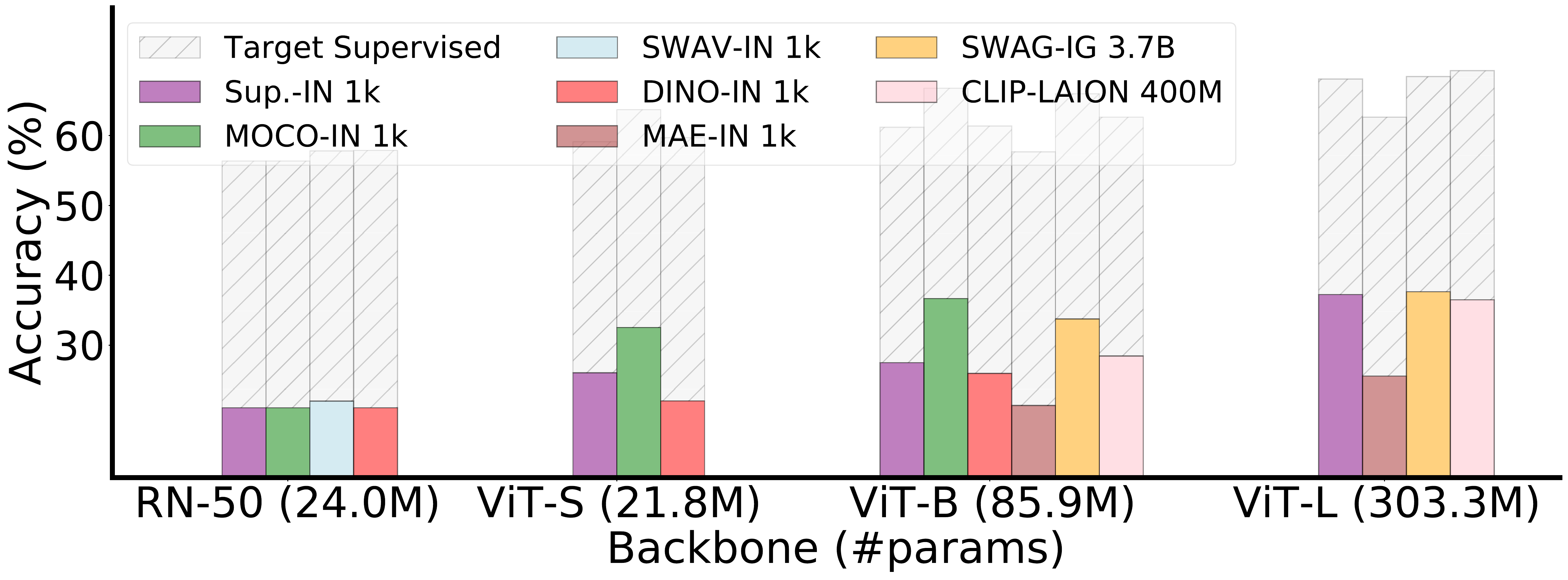}
        \vspace{-8pt}
        \subcaption{{\bf Asia$\rightarrow$USA on \GeoP{}} }
        \label{fig:vit_places_au}
    \end{minipage}
    \hfill
     \vspace{.2cm}
     \begin{minipage}[b]{0.45\textwidth}
        \centering
        \includegraphics[width=\textwidth]{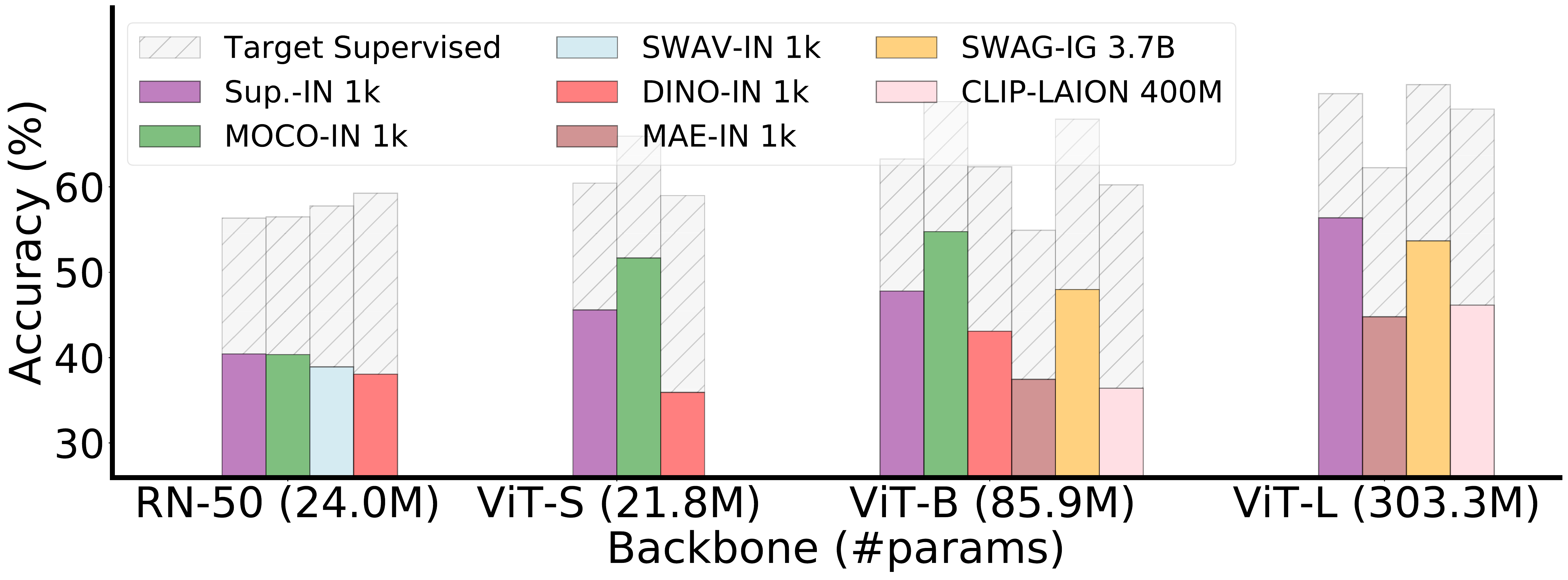}
        \vspace{-8pt}
        \subcaption{{\bf Asia$\rightarrow$USA on \GeoI{}}}
        \label{fig:vit_imnet_au}
    \end{minipage}
    \captionsetup{width=\textwidth, font=footnotesize}
    \caption{{\bf Large-Scale pre-training on \Ours{}} We show that most architectures and pre-training strategies exhibit significant cross-domain drops when fine-tuned on geographically biased datasets. Shown for Asia$\rightarrow$USA on \GeoP{} in \cref{fig:vit_places_au} and \GeoI{} in \cref{fig:vit_imnet_au}, refer main paper for other transfer settings.} 
    \label{fig:vit_asia2usa}
    \vspace{-3em}
\end{center}
\end{figure*}
%%%%%%%%%%%%%%%%%%%%%%%%%%%%%%%%%%%%%%%%%%%%%%%%%%%%%%%%%%
%%%%%%%%%%%%%%%%%%%%%%%%%%%%%%%%%%%%%%%%%%%%%%%%%%%%%%%%%%

\section{Geographic Distribution of Images}

While we broadly categorize Asia and USA to be the two major geographical domains, not all sub-regions in these geographies have equal representation. We show the geographic distribution over respective geographies in \cref{fig:worldmaps_dist}, by leveraging the per-image GPS metadata provided in \Ours{}. For images from Asia from \cref{fig:map_places_asia} for \GeoP{} and \cref{fig:map_imnet_asia} for \GeoI{},  we observe a large fraction of images from Japan, India, Korea, China and Taiwan, while some countries are more sparsely represented. Likewise, in USA in \cref{fig:map_places_usa} and \cref{fig:map_imnet_usa}, we observe a significant share of images from California, New York and Florida than other regions. These distributions reflect the larger user demographic biases in photo-sharing websites like Flickr from where all our images have been taken from.  

\section{Error Analysis of Unsupervised Adaptation}

While we show in the main paper (Table 3) that existing unsupervised adaptation approaches yield limited benefit for geographical adaptation, we conduct a deeper analysis into the per-class accuracy post-adaptation in \cref{fig:perclass_accuracies} for the case of USA$\rightarrow$Asia on \GeoP{}. Specifically, we first take a model trained only on USA images, and compute the drop in per-class accuracy suffered by direct cross-domain transfer on Asia test images. We show this in \cref{fig:da_plain}, where classes like \textit{mausoleum}, \textit{assembly line} and \textit{kitchen} suffer the largest drops in accuracy. Next, we carry the same analysis using a model trained with CDAN~\cite{CDAN} adaptation method. From \cref{fig:da_cdan}, we observe that the trends in per-class accuracy drops are mostly similar with or without using CDAN adaptation, indicating that the benefit achieved using an adaptation method is negligible on all the categories. Similar observations also hold for the case of adaptation using ToAlign~\cite{wei2021toalign}, underlining the fact that existing state-of-the-art adaptation methods cannot handle geographic shifts across most categories. 

\section{Data De-duplication}

Since a lot of users tend to upload multiple pictures of the same scene on sites like Flickr, we carry a data de-duplication exercise so that there are no such duplicate copies of same images in train and test sets which would unfairly improve within-domain accuracy. We first group all the images in the train and test sets which belong to the same geographical location, by discretizing the GPS coordinates within one degree. Then, within each group, we first resize the images to 32x32x3, and compute a histogram of the images along the RGB channels. We also flatten the image and compute the euclidean distance between all pairs of images within the same group and remove all images from the training set which are ``similar'' to images in test set, where two images are similar if they belong to the same GPS group, and have RGB histogram, euclidean distance lower than preset thresholds. 

\section{Large-scale pre-training on \Ours{}}

In \cref{fig:vit_asia2usa}, we show the effect of large-scale pretraining on the transfer setting Asia$\rightarrow$USA from \GeoP{}(\cref{fig:vit_places_au}) and \GeoI{}(\cref{fig:vit_imnet_au}). We make similar observations as the transfer setting from USA$\rightarrow$Asia in the main paper. Specifically, we show that transformers outperform Resnets, pre-training using billion-scale datasets like SWAG~\cite{singh2022revisiting} outperforms ImageNet-pretraining and all models still have significant gap with the target supervised accuracy indicating the limitations of these models in bridging cross-geography domain shifts.

\section{Effect of label-cleaning on \GeoI{}}
\begin{table}[!t]
  \centering
  \resizebox{0.46\textwidth}{!}{
  % \begin{tabular}{@{} >{\raggedright}p{5.2cm} >{\raggedright}p{2.2cm} >{\raggedright}p{2.2cm} >{\raggedright}p{2cm} p{2cm}}
  \begin{tabular}{lccccccccccc}
    & \multicolumn{6}{c}{\textbf{\underline{\GeoI}-Before Filtering}} \\
    Source $\downarrow$ / Target $\rightarrow$ & \multicolumn{2}{c}{USA} & \multicolumn{2}{c}{Asia} & \multicolumn{2}{c}{Drop(\%)} \\
    & Top-1 & Top-5 & Top-1 & Top-5 & Top-1 & Top-5 \\ 
    \cline{2-7}
    USA & 46.63 & 67.85 & 29.69 & 51.43 & \textcolor{brown}{-16.94} & \textcolor{brown}{-16.42} \\
    Asia & 31.55 & 52.28 & 52.93 & 72.96 & \textcolor{brown}{-21.38} & \textcolor{brown}{-20.68} \\
    % \textcolor{gray}{USA+Asia} & \textcolor{gray}{yy.yy/yy.yy} & \textcolor{gray}{xx.xx/xx.xx} &  \\
    \midrule
    & \multicolumn{6}{c}{\textbf{\underline{\GeoI}-After Filtering}} \\
       & \multicolumn{2}{c}{USA} & \multicolumn{2}{c}{Asia} & \multicolumn{2}{c}{Drop(\%)} \\
     & Top-1 & Top-5 & Top-1 & Top-5 & Top-1 & Top-5 \\ 
    \cline{2-7}
    USA & 56.35 & 77.95 & 36.98 & 63.42 & \textcolor{brown}{-19.37} & \textcolor{brown}{-14.53} \\
    Asia & 40.43 & 64.60 & 60.37 & 80.22 & \textcolor{brown}{-19.94} & \textcolor{brown}{-15.62} \\
    % \textcolor{gray}{USA+Asia} & \textcolor{gray}{yy.yy/yy.yy} & \textcolor{gray}{xx.xx/xx.xx} &  \\
    \bottomrule
  \end{tabular}
  }
  \vspace{-1em}
  \captionsetup{width=0.45\textwidth}
  \caption{\label{tab:prefilter} Top-1/Top-5 accuracy of models across geographically different train and test domains with a more noisier 700-class version of \GeoI{}.}
  \vspace{-1em}
\end{table}

Before the current version of \GeoI{} with 600 classes, we created a slightly larger, albeit more noisy 700 class version. We then observed that although these concepts have been selected from ImageNet, there were many ambiguous classes (like \texttt{fancy dress, frozen yogurt, prey, flash, walking stick}) etc. So, we removed 100 such classes with ambiguous concept meanings, and created a newer version with 600 classes, which is eventually used in benchmarking and release. In \cref{tab:prefilter}, we show the cross-domain accuracies with the older version. We observed that while the cross-domain drops remain the same, the absolute accuracy themselves are much higher using a cleaner version of the dataset. 

% \section{Dataset Details}

% The category list in Places-205 dataset, which we also use in our \GeoP{} benchmark, along with the list of concepts used in our \GeoI{} benchmark, which are a strict subset of the ILSVRC-21 dataset~\cite{deng2009imagenet} are provided as attachments along with the supplementary pdf material.

\section{Sample Images} 
\begin{figure*}[!thbp]
    \begin{center}
    
    \begin{subfigure}[b]{\textwidth}
        \centering
        \includegraphics[width=\textwidth]{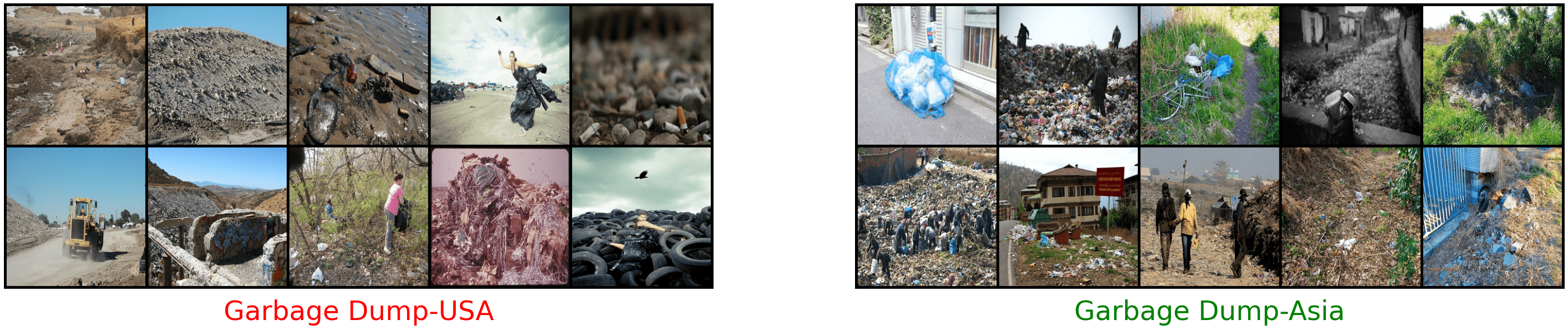}
    \end{subfigure}
    \vspace{10pt}
    \begin{subfigure}[b]{\textwidth}
        \centering
        \includegraphics[width=\textwidth]{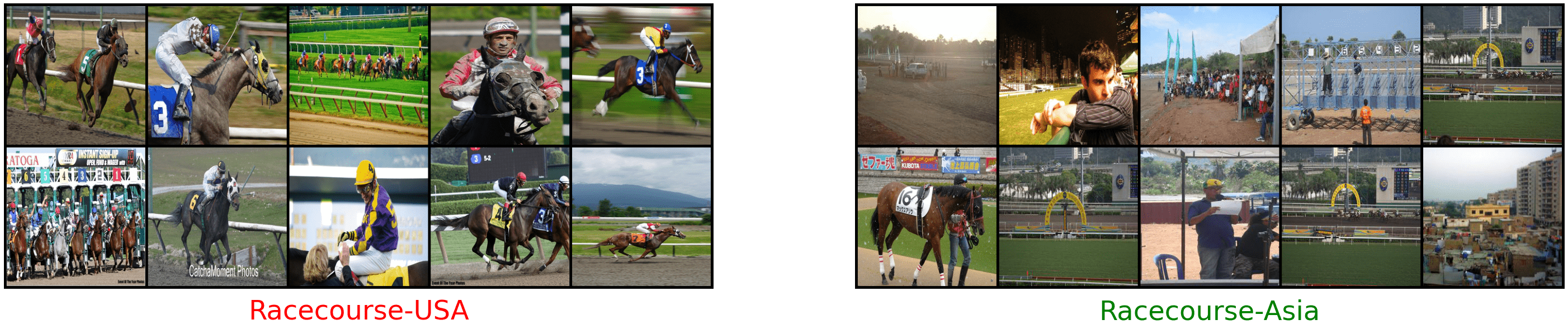}
    \end{subfigure}
    \vspace{10pt}
    \begin{subfigure}[b]{\textwidth}
        \centering
        \includegraphics[width=\textwidth]{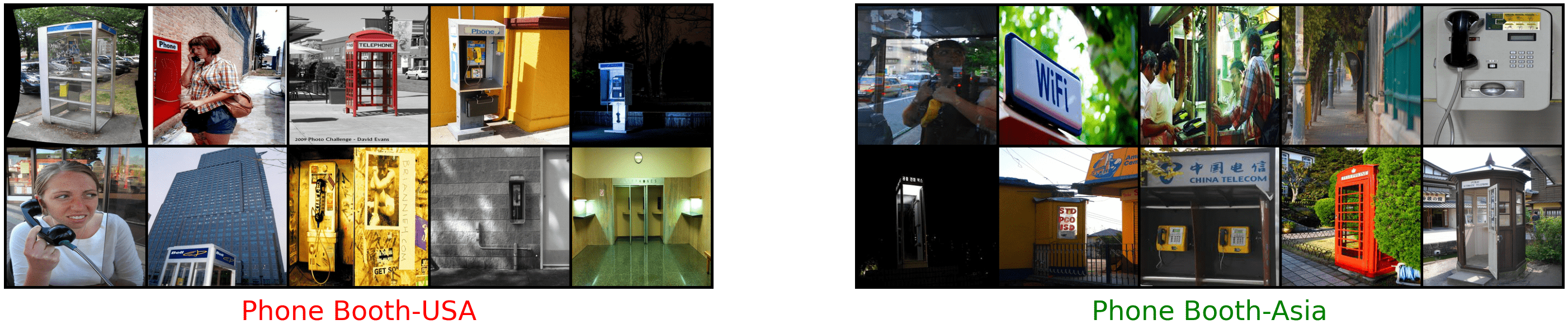}
    \end{subfigure}
    \vspace{10pt}
    \begin{subfigure}[b]{\textwidth}
        \centering
        \includegraphics[width=\textwidth]{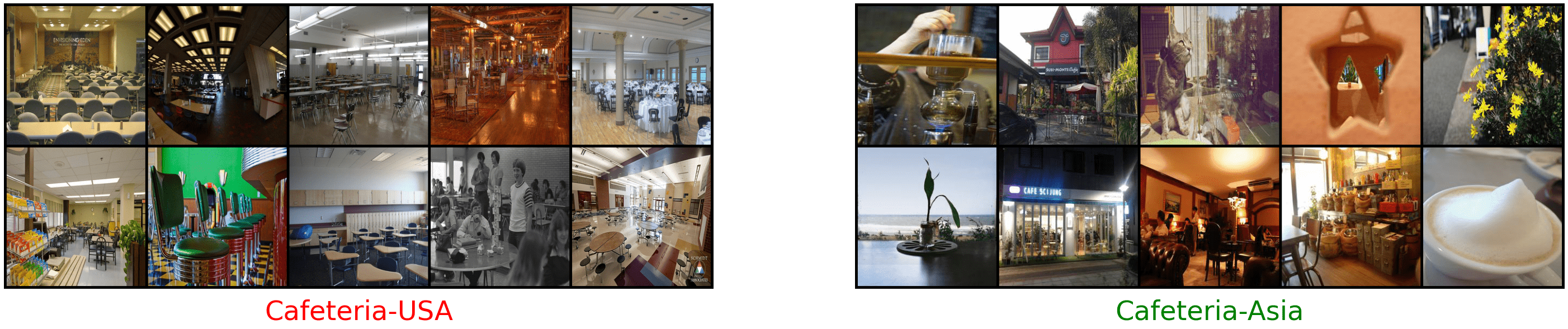}
    \end{subfigure}
    %%%%%%%%%%%%%%%%
    %%%%%%%%%%%%%%%%
    
    \end{center}
    \captionsetup{width=0.8\textwidth}
    \caption{Sample images showing the domain gap between USA (left) and Asia (right) domains for classes \texttt{garbage dump}, \texttt{ race course}, \texttt{ phone booth} and \texttt{ cafetaria} from \GeoP{}.}
    \label{fig:sample_images_1}
\end{figure*}

\begin{figure*}[!thbp]
    \begin{center}
    
    \begin{subfigure}[b]{\textwidth}
        \centering
        \includegraphics[width=\textwidth]{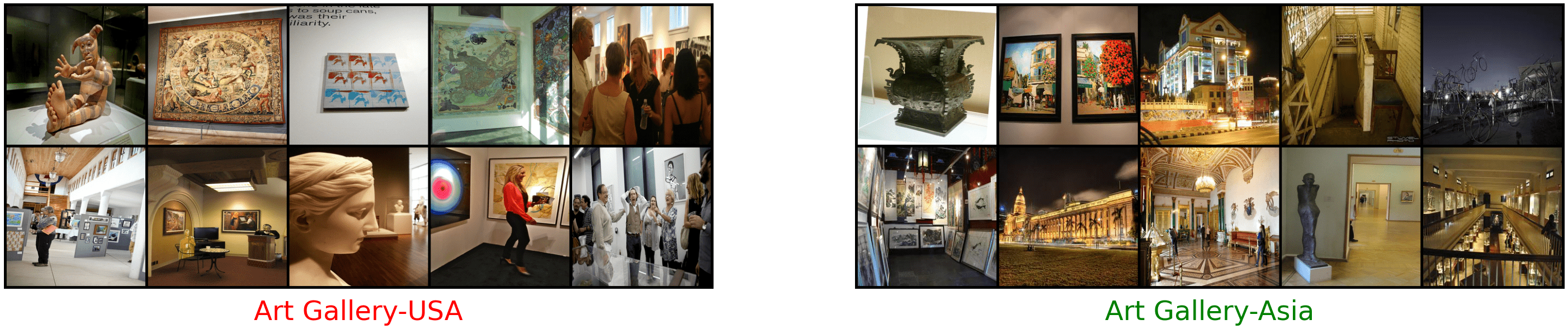}
    \end{subfigure}
    \vspace{10pt}
    \begin{subfigure}[b]{\textwidth}
        \centering
        \includegraphics[width=\textwidth]{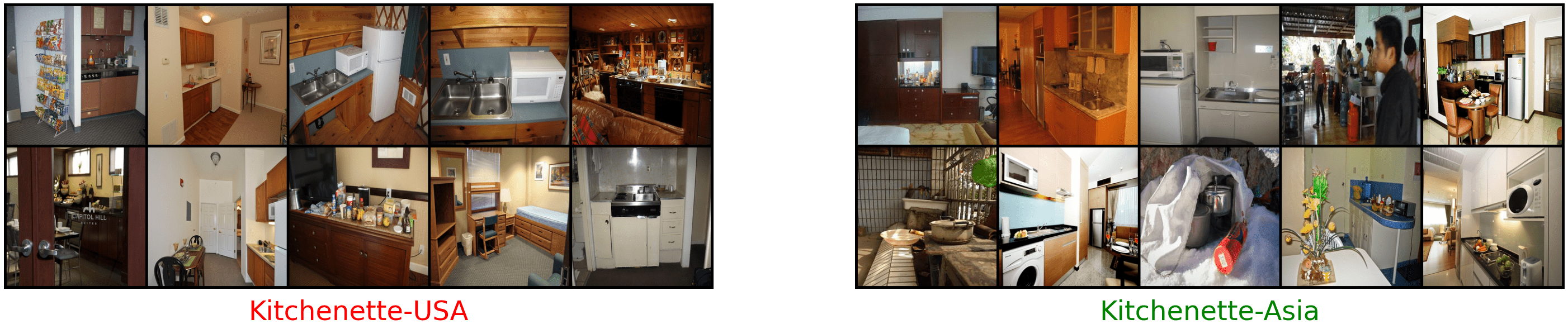}
    \end{subfigure}
    \vspace{10pt}
    \begin{subfigure}[b]{\textwidth}
        \centering
        \includegraphics[width=\textwidth]{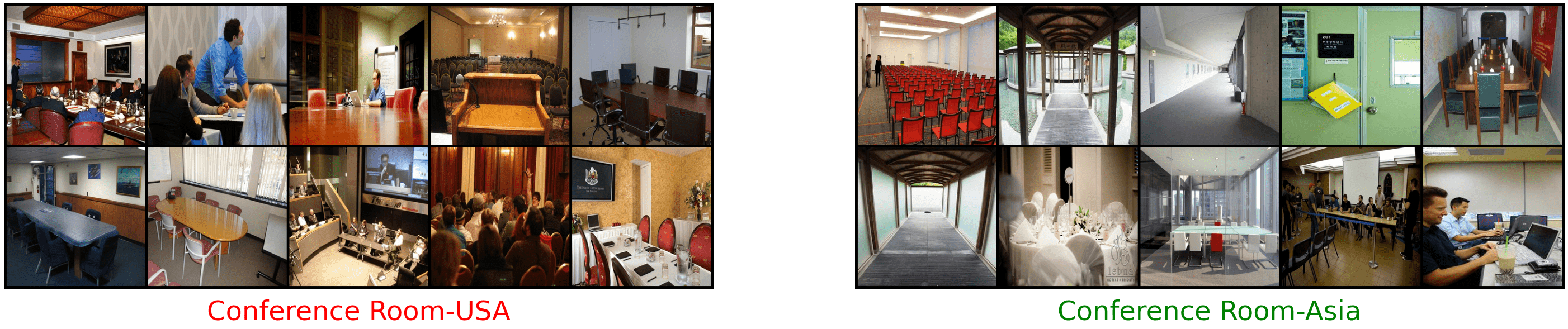}
    \end{subfigure}
    \vspace{10pt}
    \begin{subfigure}[b]{\textwidth}
        \centering
        \includegraphics[width=\textwidth]{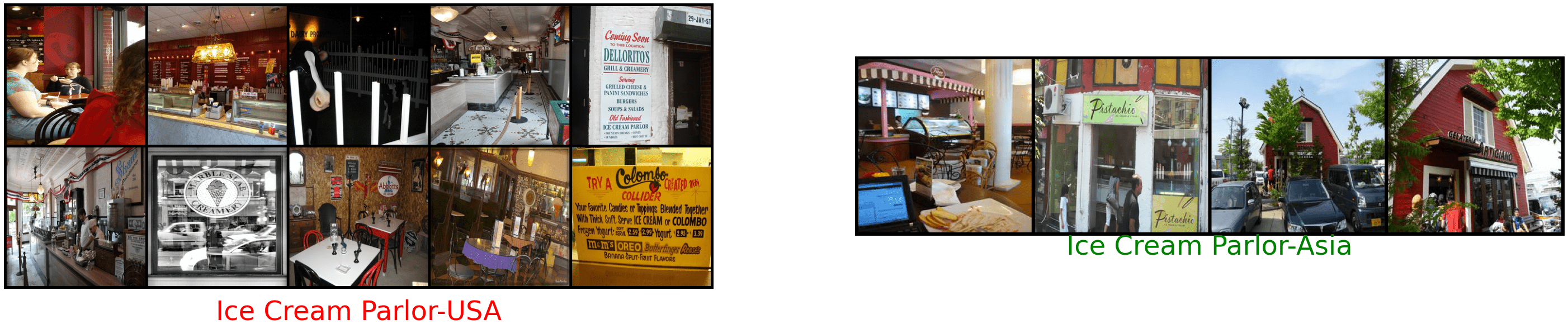}
    \end{subfigure}
    %%%%%%%%%%%%%%%%
    %%%%%%%%%%%%%%%%
    
    \end{center}
    \caption{Sample images showing the domain gap between USA (left) and Asia (right) domains for classes \texttt{art gallery}, \texttt{ kitchenette}, \texttt{ conference room} and \texttt{ ice-cream parlor} from \GeoP{}.}
    \label{fig:sample_images_2}
\end{figure*}

\begin{figure*}[!thbp]
    \begin{center}
    
    \begin{subfigure}[b]{\textwidth}
        \centering
        \includegraphics[width=\textwidth]{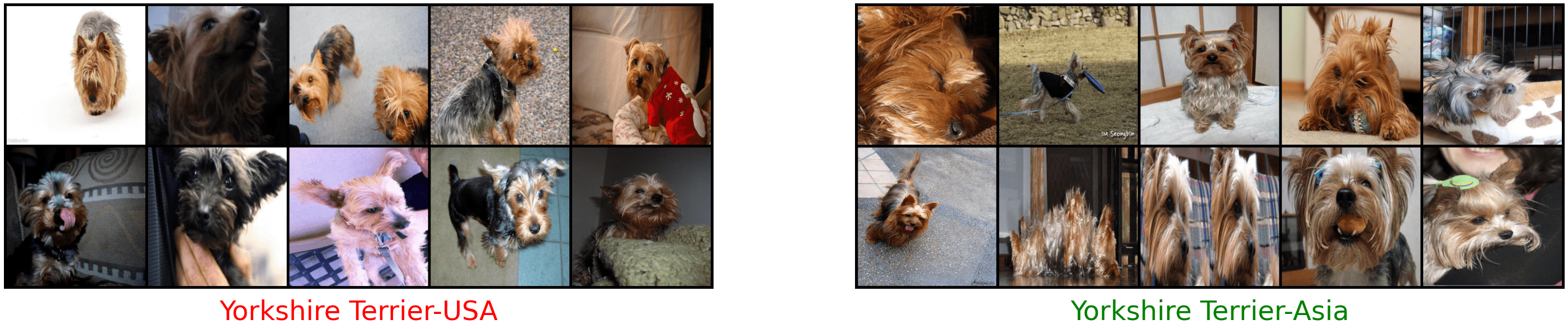}
    \end{subfigure}
    \vspace{10pt}
    \begin{subfigure}[b]{\textwidth}
        \centering
        \includegraphics[width=\textwidth]{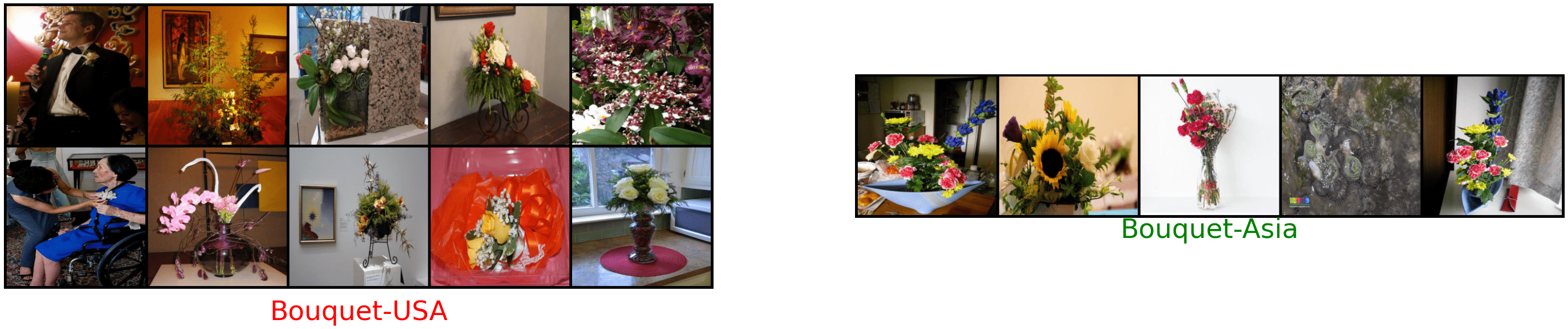}
    \end{subfigure}
    \vspace{10pt}
    \begin{subfigure}[b]{\textwidth}
        \centering
        \includegraphics[width=\textwidth]{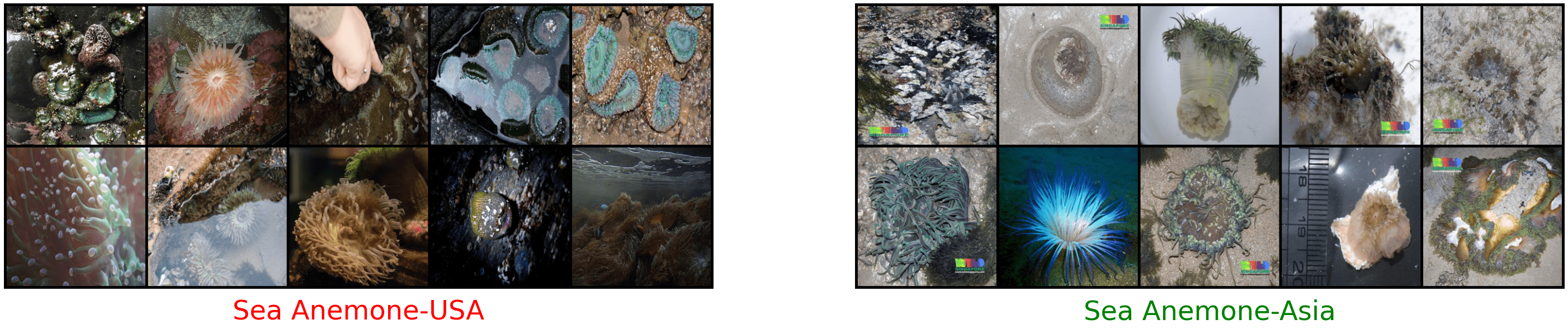}
    \end{subfigure}
    \vspace{10pt}
    \begin{subfigure}[b]{\textwidth}
        \centering
        \includegraphics[width=\textwidth]{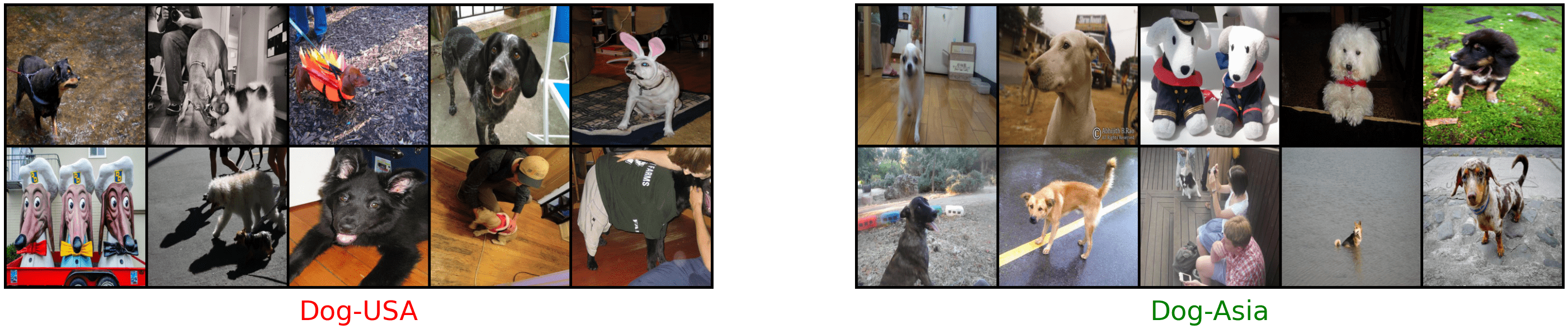}
    \end{subfigure}
    %%%%%%%%%%%%%%%%
    %%%%%%%%%%%%%%%%
    
    \end{center}
    \caption{Sample images showing the domain gap between USA (left) and Asia (right) domains for classes \texttt{Yorkshire Terrier}, \texttt{ bouquet}, \texttt{ sea anemone} and \texttt{ dog} from \GeoI{}.}
    \label{fig:sample_images_3}
\end{figure*}

\begin{figure*}[!thbp]
    \begin{center}
    
    \begin{subfigure}[b]{\textwidth}
        \centering
        \includegraphics[width=\textwidth]{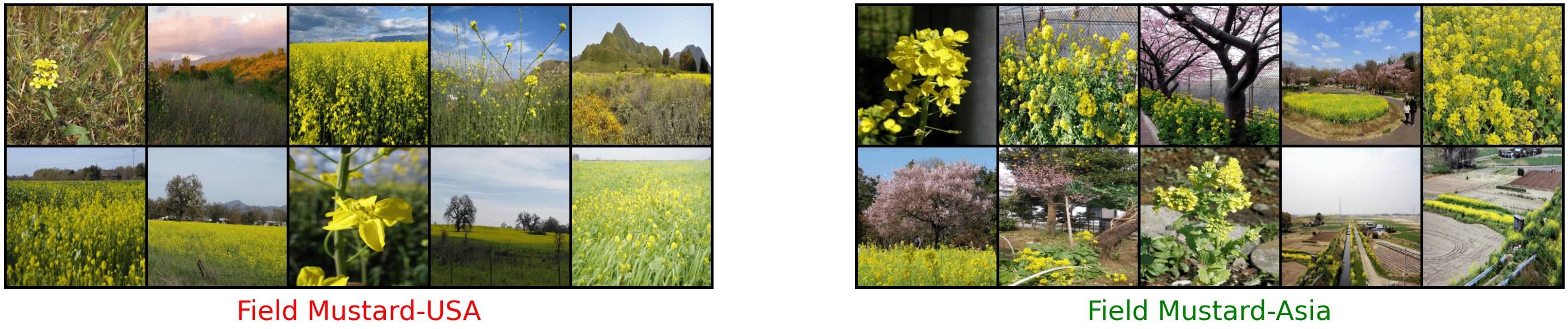}
    \end{subfigure}
    \vspace{10pt}
    \begin{subfigure}[b]{\textwidth}
        \centering
        \includegraphics[width=\textwidth]{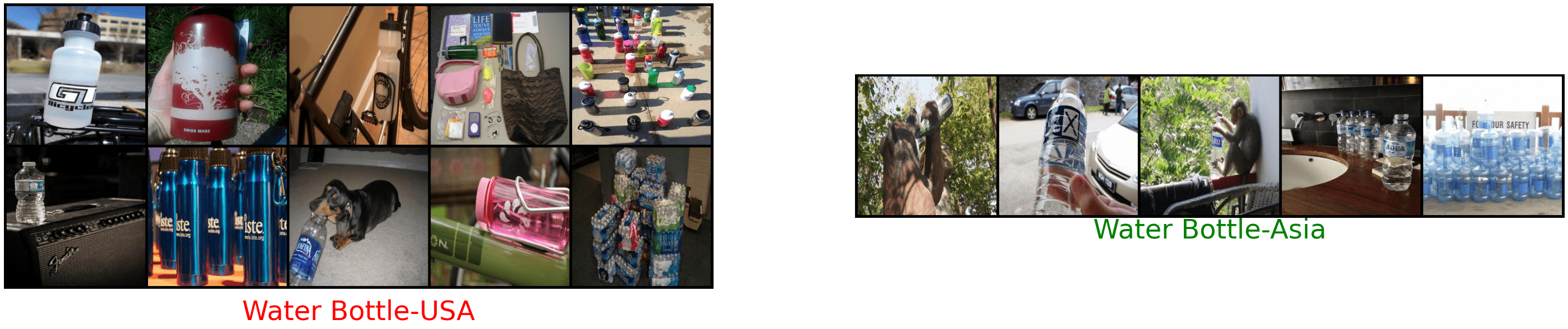}
    \end{subfigure}
    \vspace{10pt}
    \begin{subfigure}[b]{\textwidth}
        \centering
        \includegraphics[width=\textwidth]{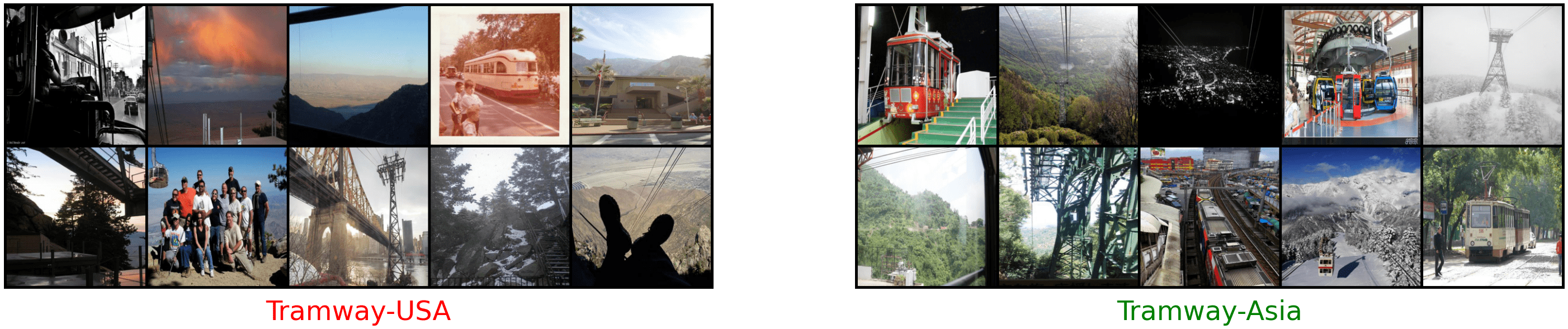}
    \end{subfigure}
    \vspace{10pt}
    \begin{subfigure}[b]{\textwidth}
        \centering
        \includegraphics[width=\textwidth]{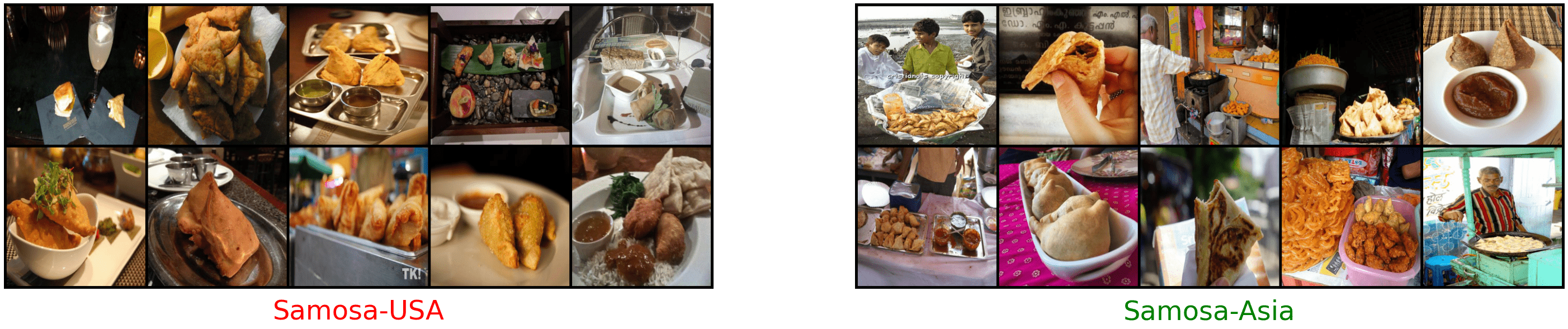}
    \end{subfigure}
    %%%%%%%%%%%%%%%%
    %%%%%%%%%%%%%%%%
    
    \end{center}
    \caption{Sample images showing the domain gap between USA (left) and Asia (right) domains for classes \texttt{Field Mustard}, \texttt{ Water Bottle}, \texttt{ Tramway} and \texttt{ Samosa} from \GeoI{}.}
    \label{fig:sample_images_4}
\end{figure*}

We show few sample images from selected classes across both USA and Asia domains in \GeoP{} benchmark in \cref{fig:sample_images_1}, \cref{fig:sample_images_2} and \GeoI{} benchmark in \cref{fig:sample_images_3}, \cref{fig:sample_images_4}.

%% Qualitative Figures

\clearpage

%%%%%%%%%%%%%%%%%%%%%%%%%%%%%%%%%%%%%%%%%%%%%%%%%%%%%%%%%%%%%
%%%%%%%%%%%%%%%%%%%%%%%%%%%%%%%%%%%%%%%%%%%%%%%%%%%%%%%%%%%%%%

%% Qualitative Figures

\clearpage

%%%%%%%%%%%%%%%%%%%%%%%%%%%%%%%%%%%%%%%%%%%%%%%%%%%%%%%%%%%%%%%%%%
%%%%%%%%%%%%%%%%%%%%%%%%%%%%%%%%%%%%%%%%%%%%%%%%%%%%%%%%%%%%%%%%%%

%%%%%%%%%%%%%%%%%%%%%%%%%%%%%%%%%%%%%%%%%%%%%%%%%%%%%%%%%%%%%
%%%%%%%%%%%%%%%%%%%%%%%%%%%%%%%%%%%%%%%%%%%%%%%%%%%%%%%%%%%%%%

%% Qualitative Figures

\clearpage

%%%%%%%%%%%%%%%%%%%%%%%%%%%%%%%%%%%%%%%%%%%%%%%%%%%%%%%%%%%%%%%%%%
%%%%%%%%%%%%%%%%%%%%%%%%%%%%%%%%%%%%%%%%%%%%%%%%%%%%%%%%%%%%%%%%%%

%%%%%%%%%%%%%%%%%%%%%%%%%%%%%%%%%%%%%%%%%%%%%%%%%%%%%%%%%%%%%
%%%%%%%%%%%%%%%%%%%%%%%%%%%%%%%%%%%%%%%%%%%%%%%%%%%%%%%%%%%%%%

%% Qualitative Figures

\clearpage

%%%%%%%%%%%%%%%%%%%%%%%%%%%%%%%%%%%%%%%%%%%%%%%%%%%%%%%%%%%%%%%%%%
%%%%%%%%%%%%%%%%%%%%%%%%%%%%%%%%%%%%%%%%%%%%%%%%%%%%%%%%%%%%%%%%%%

\clearpage
\end{appendix}

\end{document}